%% file: main.tex
\documentclass[final,5p,times,twocolumn,authoryear]{elsarticle}

\usepackage{amssymb}
\usepackage{lipsum}
\usepackage[utf8]{inputenc}
\usepackage{multirow}
\usepackage{algorithmic}
\usepackage{orcidlink}
\usepackage{subfigure}
\usepackage{subcaption}
\usepackage{amsmath} 
\usepackage{amsfonts,amssymb}  
\usepackage{colortbl}
\usepackage{amssymb}
\usepackage{bbding}
\usepackage{makecell}
\usepackage{xcolor}
\usepackage{color}
\usepackage{threeparttable} 
\usepackage{xcolor}
\usepackage{graphicx}
\usepackage{mathrsfs}
\usepackage{stfloats}
\usepackage{algorithm,algorithmic}
\graphicspath{ {./figures/} }
\definecolor{mycolor}{RGB}{188,231,204}
\definecolor{mycolor2}{RGB}{228,238,188}
\definecolor{mycolor3}{RGB}{254, 248, 198}

\begin{document}

\begin{frontmatter}



\title{Diffusion Posterior Sampler for Hyperspectral Unmixing with Spectral Variability Modeling}


\author{Yimin Zhu, Lincoln Linlin Xu}
\affiliation{organization={Department of Geomatics Engineering, University of Calgary, Calgary, Canada}}

\begin{abstract}
Linear spectral mixture models (LMM) provide a concise form to disentangle the constituent materials (endmembers) and their corresponding proportions (abundance) in a single pixel. The critical challenges are how to model the spectral prior distribution and spectral variability. Prior knowledge and spectral variability can be rigorously modeled under the Bayesian framework, where posterior estimation of Abundance is derived by combining observed data with endmember prior distribution. Considering the key challenges and the advantages of the Bayesian framework, a novel method using a diffusion posterior sampler for semiblind unmixing, denoted as DPS4Un, is proposed to deal with these challenges with the following features: (1) we view the pretrained conditional spectrum diffusion model as a posterior sampler, which can combine the learned endmember prior with observation to get the refined abundance distribution. (2) Instead of using the existing spectral library as prior, which may raise bias, we establish the image-based endmember bundles within superpixels, which are used to train the endmember prior learner with diffusion model. Superpixels make sure the sub-scene is more homogeneous. (3) Instead of using the image-level data consistency constraint, the superpixel-based data fidelity term is proposed. (4) The endmember is initialized as Gaussian noise for each superpixel region, DPS4Un iteratively updates the abundance and endmember, contributing to spectral variability modeling. The experimental results on three real-world benchmark datasets demonstrate that DPS4Un outperforms the state-of-the-art hyperspectral unmixing methods.
\end{abstract}

\end{frontmatter}




\input{0_abstract}    
\input{1_intro}

\input{2_Related_work}

\input{3_Methodology}
\input{4_Results_and_Analysis}

\input{5_Conclusion}

\bibliographystyle{elsarticle-harv} 
\bibliography{example}

\end{document}

%% file: 0_abstract.tex
\begin{abstract}
Linear spectral mixture models (LMM) provide a concise form to disentangle the constituent materials (endmembers) and their corresponding proportions (abundance) in a single pixel. The critical challenges are how to model the spectral prior distribution and spectral variability. Prior knowledge and spectral variability can be rigorously modeled under the Bayesian framework, where posterior estimation of Abundance is derived by combining observed data with endmember prior distribution. Considering the key challenges and the advantages of the Bayesian framework, a novel method using a diffusion posterior sampler for semiblind unmixing, denoted as DPS4Un, is proposed to deal with these challenges with the following features: (1) we view the pretrained conditional spectrum diffusion model as a posterior sampler, which can combine the learned endmember prior with observation to get the refined abundance distribution. (2) Instead of using the existing spectral library as prior, which may raise bias, we establish the image-based endmember bundles within superpixels, which are used to train the endmember prior learner with diffusion model. Superpixels make sure the sub-scene is more homogeneous. (3) Instead of using the image-level data consistency constraint, the superpixel-based data fidelity term is proposed. (4) The endmember is initialized as Gaussian noise for each superpixel region, DPS4Un iteratively updates the abundance and endmember, contributing to spectral variability modeling. The experimental results on three real-world benchmark datasets demonstrate that DPS4Un outperforms the state-of-the-art hyperspectral unmixing methods.
\end{abstract}

%% file: 1_intro.tex
\section{Introduction}
\label{sec:intro}
\begin{figure}[!h]
    \centering
    \includegraphics[width=1\linewidth]{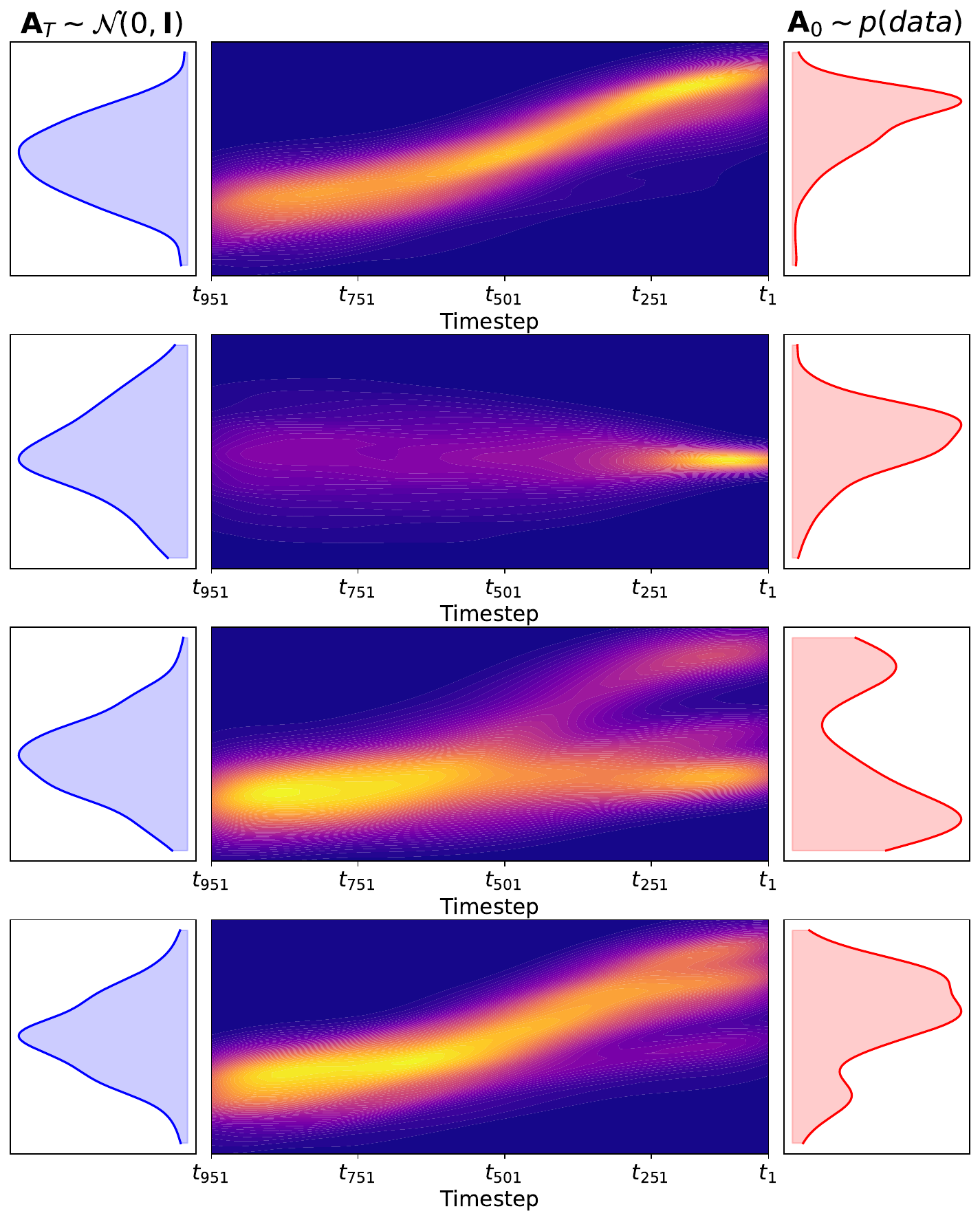}
    \caption{Sampling trajectory of our proposed DPS4Un from Gaussian noise (left) to endmember distribution (right) considering the spectral variability (multiple trajectories). The \(x\)-axis represents the DDIM sampling timestep \(t\), total 20 steps, the \(y\)-axis represents the value of the denoised endmember \(\boldsymbol{A}_t\). The color is the probability density of that value at that timestep. Overall, the reverse process should flow toward high probability regions of the data distribution, showing that our proposed DPS4Un can path directly to the high-mass target region, which is good for the spectral unmixing problem. To show the high-dimensional data, PCA is applied to reduce the data into a 2D space.}
    \label{figure1}
\end{figure}
Hyperspectral images (HSI) offer rich spatial-spectral information that can help to distinguish various spectrally-similar objects to support various environmental applications \cite{PAOLETTI2019279}. Due to the rich spectral information, making hyperspectral image be used in various socio-environmental applications, including atmospheric environment monitoring \cite{Ghamisi2017, Alakian2024, Rouet-Leduc2024}, geologic mapping \cite{HAJAJ2024101218, Acosta2019}, agriculture disaster response \cite{AKHYAR2024112067, ZHENG2021112636}, and medical and surgical diagnostics \cite{medical_hyper}. However, due to the requirement of signal-to-noise ratio and the field of view required, the spatial resolution of hyperspectral images is usually coarse, making a pixel of hyperspectral data usually include multiple pure materials (mixed pixels), and this spectral mixture greatly impairs the performance of extracting information from the data. Spectral unmixing (SU) models aim to decompose the mixed pixels \(\mathbf{X}\) to derive both the spectral signatures of constituent components (i.e., endmembers, \(\mathbf{A}\)) and their corresponding fractional proportions (i.e., abundances, \(\mathbf{S}\)) \cite{6200362}. Two parameters need to be solved with only one observation, making SU a challenging ill-posed inverse problem \cite{6200362}. Effective SU approaches rely on accurately modeling the \textit{endmember prior}, \textit{spectral variability}, and an effective \textit{iterative optimization} method to estimate endmember and abundance.

However, the existing SU methods have some limitations in spectral decomposition solving. First, although the United States Geological Survey (USGS) spectral library provides valuable spectral prior information for enhanced SU performance \cite{iordache2011sparse}, but existing semiblind SU approaches are still limited by the model's nonconvex properties, resulting in model sensitivity to the regularization parameter \cite{iordache2012total, wei2022multiobjective}. Endmember bundles extraction is a more efficient approach for building spectral prior by considering the regional spatial and spectral closeness \cite{xu2018regional, shi2022deep}. Although this method can provide rich endmember candidates, but it still serves as preprocessing tools, and the spectral prior can not be updated by some learning approaches. Therefore, the essential challenge lies in how to learn a more comprehensive spectral prior that captures both global spectral characteristics and local variability in a data-driven manner.

Second, spectral-spatial heterogeneity reflected by the trade-off on spectral and spatial resolution \cite{zhong2020whu}, low signal-to-noise ratio, as well as lighting effect \cite{9583297}, leading to significant spectral variability \cite{han2025subpixel, 9439249}. Each pixel should have its own unique endmembers \cite{9439249}. Traditional spectral unmixing (SU) algorithms, such as N-FINDR \cite{winter1999n}, Vertex Component Analysis (VCA) \cite{nascimento2005vertex}, and Pixel Purity Index (PPI) \cite{chang2006fast}, neglect the spectral variability of the endmembers, which propagates significant modeling errors throughout the whole unmixing process and compromises the quality of the results. Some autoencoder-based networks for SU use VCA endmember \cite{9709848} or random \cite{10902420} initialization or the weights of the decoder \cite{9848995, zeng2024unmixing} for endmember update. However, these models are more focused on the spatial correlation learning through FCNN-like architecture, instead of the endmember variability modeling. Therefore, for better abundance estimation, it is necessary to consider the spectral variability modeling.

Third, considering the above-mentioned two key factors, once the spectral prior is built, the SU process can be generalized as a maximum a posteriori problem in a Bayesian framework. Utlizing \(p(\mathbf{A})\) as prior, and interatively samples from the posterior \(p(\mathbf{A}, \mathbf{S}|\mathbf{X})\) established by \(p(\mathbf{A}, \mathbf{S}|\mathbf{X})=p(\mathbf{X}|\mathbf{A},\mathbf{S})p(\mathbf{S}|\mathbf{A})p(\mathbf{A})/p(\mathbf{X})\). This iterative method allows the estimated abundances and endmembers to be updated and refined in an adaptive manner. Hence, building a pipeline for interactive sampling for the SU problem needs to be researched.

In recent years, generative models represented by diffusion models have been proven that it can used to solve linear (e.g., image super-resolution, image restoration, Gaussian deblur, inpainting, motion deblur) and non-linear (e.g., phase retrieval, non-uniform deblur) reverse problems \cite{chung2023diffusion, daras2024survey, song2021solving}. Motivated by the data distribution modeling capability of the diffusion model, we formulate the spectral prior modeling as a conditional spectrum generative task. In the prior learning stage, we employ the diffusion model conditioning on the endmember bundle ID for generating the specific spectrum while preserving detailed information. At the sampling stage, an interactive sampling method is designed to circumvent the intractability of posterior sampling by the conditional diffusion model via a novel approximation. The main contributions are threefold:

\begin{itemize}

    \item We rethink the semiblind SU problem from the perspective of the generative tasks, and decouple this question into endmember prior learning and posterior sampling process to refine the endmember and abundance. Instead of using the unconditional diffusion model, we treat the region-based clustering ID as conditions that allow the model to generate more reasonable spectrum with less randomness. This idea is completely different from other semiblind unmixing methods due to its new probabilistic distribution estimation, inference, and sampling.
    
    \item In order to model the spectral variability, instead of starting from Gaussian noise for global endmember denoising, we initialize for each superpixel region to better capture the spectral variability. Additionally, the loss between data and likelihood is also region-based for solving the spatial smoothness.

    \item The estimation of abundance and endmember totally performs posterior sampling using the pretrained diffusion model, which serves as the data prior.

\end{itemize}

%% file: 2_Related_work.tex
\section{Related Works}
\label{sec:Related Works}

\subsection{Hyperspectral unmixing}

\paragraph{Linear Spectral Mixture Model.} Let assume that HSI cude with total \(C\) spectral bands and \(N\) pixels, each pixel at site \(i\) denoted by \(y_i\) being a \(C \times 1\) vector. The HSI is assumed to contain \(K\) endmembers. According to Linear Spectral Mixture Model (LSMM), the HSI cube \(\boldsymbol{X} \in \mathbb{R}^{C \times N}\) is represented by the product of the endmember matrix \(\boldsymbol{A} \in \mathbb{R}^{C \times K}\) and the abundance matrix \(\boldsymbol{S} \in \mathbb{R}^{K \times N}\), with some additive Gaussion noise \(\boldsymbol{N} \in \mathbb{R}^{C \times N}\), as follows:
\begin{align}
    \begin{aligned}
        &\boldsymbol{X} = \boldsymbol{A} \boldsymbol{S} + \boldsymbol{N} \\
        s.t. \quad &\underset{k}{\sum} \boldsymbol{s}_{i}^{k}=1, \boldsymbol{s}_{i}^k \ge 0, i=1,\dots,N
    \end{aligned}
\end{align}
where, \(s_i^k\) is the abundance value at site \(i\) for \(k-th\) endmember. The noise term \(\boldsymbol{N}\) distribution is assumed to satisfy a Gaussian distribution model by \(\boldsymbol{N}_{ci} = \mathcal{N}(0, \sigma_{c}^2)\), where \(\sigma_{c}^2\) is noise variance of the each band.

The hyperspectral unmixing problem modeled by LSMM can
be considered as an inverse problem to restore the \(\boldsymbol{A}\) and \(\boldsymbol{S}\) from \(\boldsymbol{X}\), which in a Bayesian framework, can be achieved by maximizing the posterior distribution \(p(\boldsymbol{A},\boldsymbol{S}|\boldsymbol{X})\), i.e.,
\begin{align}
    \begin{aligned}
        p(\boldsymbol{A},\boldsymbol{S}|\boldsymbol{X}) \propto p(\boldsymbol{X}|\boldsymbol{A}, \boldsymbol{S})p(\boldsymbol{S}|\boldsymbol{A})p(\boldsymbol{A})
    \end{aligned}
\end{align}

\paragraph{Semiblind Unmixing.} Semiblind unmixing approaches are mainly sparse regression methods that select a small subset of endmembers from a given large-scale spectral library \cite{iordache2011sparse}. The algorithms with spectral dictionary-aided models treat the SU problem as a single-pixel-based sparse regression problem and a multiple-pixel-based collaborative sparse problem, respectively. \(\ell_1\) or mixed \(\ell_1 / \ell_2\) optimization methods are used to measure the sparsity of the abundance matrix \cite{iordache2012total, iordache2013collaborative, shi2018collaborative}. However, high mutual coherence of the large-scale spectral library may lead
to poor performance when applying the above minimization-based sparse optimization \cite{fu2016semiblind}. Some Bayesian inference models, such as \cite{9709848}, use deep learning prior (DIP) to estimate the abundance, but this method relies on the availability of an estimation of the endmembers using other existing methods. In order to provided multiple endmembers prior for semiblind SU, \cite{shi2022deep} used extracted endmember bundles to constrain the generator training with GAN training style.

\paragraph{Blind Unmixing.} Blind unmixing methods determine both the endmembers and their abundances simultaneously, which can be formulated as a nonnegative
matrix factorization (NMF) problem. Traditional methods, such as \cite{fevotte2015nonlinear, feng2018hyperspectral, miao2007endmember}, decompose a nonnegative observations matrix into two nonnegative matrices. Some autoencoder-based (AE) networks are proposed to solve the SU problem, where the bottleneck of the AE provides the abundance estimation, and the weights of the decoder provide the endmember estimation \cite{qu2018udas, ozkan2018endnet, su2019daen}. Some models, such as \cite{han2025subpixel, fang2024hyperspectral}, extend the decoder part to solve the multilinear or nonlinear SU problem. Double DIP models \cite{zhou2023hyperspectral, zhou2022blind} are also proposed for blind unmixing; however, the spectral variability hasn't been considered. In order to model the spectral variability, the latent variable models, such as variational AE (VAE), is introduced to control the variability of endmembers by using Gaussian noise to represent the posterior distribution of endmember \cite{shi2021probabilistic, shi2021variational}.

\subsection{Diffusion model for inverse problem} \label{inverse problem}

Diffusion model learn the implicit prior of the underlying data distribution by matching the gradient of the log density \(\nabla_x \log p(x)\). This prior can be leveraged to solve the inverse problem, aiming to recover \(x\) from observation \(y\). While, due to the intractable likelihood \(\nabla_x \log p(y|x)\) in the diffusion model, many projection-based methods are proposed to solve it \cite{liu2023accelerating, choi2021ilvr, chung2022improving}. For yielding approximate posterior sampling, the Tweedie formula is used to provide a tractable approximation for \(p(y|x_t)\) using \(p(y|\hat{x}_0)\) instead. With this approximation, different noise terms (Gaussian noise or Poisson noise) in the forward model can be considered in the diffusion posterior sampling \cite{chung2023diffusion}. Hence, more general noisy inverse problems can be solved. However, these methods need to know the operator in the forward model. Hence, BlindDPS, is proposed to also estimate the parameter of the forward model \cite{chung2023parallel}.

In the field of remote sensing, diffusion models are also used for super-resolution \cite{xiao2023ediffsr, han2023enhancing, meng2024conditional, 10644098} and denoising \cite{Zeng_2024_CVPR, perera2023sar}. \cite{10644098} leveraged the diffusion model to capture the spectral distribution knowledge and accordingly exploit a spectral-level prior
to guide the fusion of low-resolution HSI with high-resolution multispectral image, showing accurate spectrum generation and fusion capability. 

However, A few of the works plug the diffusion prior for the hyperspectral unmixing task. From our knowledge, only one recent work \cite{deng2024diffusion} has been proposed to solve the SU problem. While our main differences with \cite{deng2024diffusion} are that our proposed method is a conditional diffusion model, which can generate accurate spectra under the control of spectral categorical ID, while, \cite{deng2024diffusion} is unconditional, which may lead to random spectral generation and inaccurate estimation of abundance \(\boldsymbol{S}\). Additionally, \cite{deng2024diffusion} didn't consider the spectral variability.

%% file: 3_Methodology.tex
\section{Methodology}
\label{sec:methods}

\subsection{Main Idea} Some existing methods that plug the diffusion model are basically unconditional ones \cite{deng2024diffusion, 10644098}, while, due to the randomness of the diffusion reverse process, without any guidance or conditions, they can not make sure that the trajectory of the diffusion sampling process will arrive at the desired endmember's distribution. Another thing is the consideration of the spectral variability quantification, which is 
also the approach \cite{deng2024diffusion} lacks. Variability quantification is important for the diffusion ensemble models in environmental and remote sensing topics \cite{li2024generative, 11206357, yu2025probabilistic, finn2024generative} 

To better solve these problems, in this section, we demonstrated that our proposed conditional spectrum diffusion prior learner (log density, \(\nabla _x \log p(x_t|c)\), \(c\) is the condition) can provide a more realistic sampling trajectory from Gaussian noise to the desired endmember's distribution, which reduces the randomness of the model. Moreover, since our explicit approximations for the measurements matching term \(\nabla _{\boldsymbol{A}_t} \log p(\boldsymbol{X}|\hat{\boldsymbol{A}}_0, \hat{\boldsymbol{S}}_{0}, c)\) is region (superpixel) based, each region will be initialized Gaussian noise during sampling stage, making sure that each region have their own different our endmember sets, contributing to spectral variability modeling, instead of a global endmember for the entire hyperspectral image. \(\hat{\boldsymbol{A}}_0\) is the approximation for the posterior mean, as we talked in \autoref{inverse problem}. \(\hat{\boldsymbol{S}}_{0}\) is the optimal estimation of the abundance matrix using projected gradient descent given \(\hat{\boldsymbol{A}}_0\) and \(\boldsymbol{X}\).

\subsection{Superpixel-based Spectral Library}

\begin{figure}
    \centering
    \includegraphics[width=0.49\textwidth]{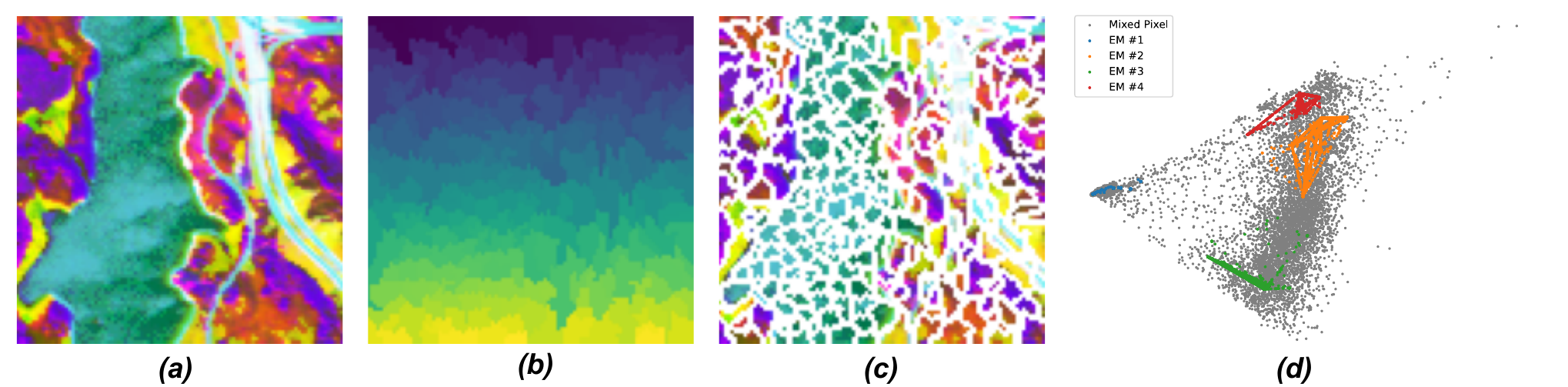}
    \caption{Illustration for region-based spectral library construction. (a) HSI in HSV color space. (b) superpixel (c) superpixel overlay on HSI. (d) Data distribution of mixed pixels (grey) and spectral library (colored) in 2D PCA space.}
    \label{spectral library}
\end{figure}

Instead of using a large-scale library from USGS, we utilize a superpixel-based endmember bundle extraction method in \cite{xu2018regional} to build
a specific image-level spectral library. Specifically, we split the HSI into \(L\) subsets using the Simple Linear Iterative Clustering (SLIC) Algorithm \cite{achanta2012slic} for superpixel generation, which jointly considers the spatial Euclidean distance and spectral similarity, as follows: 
\begin{align}
    \begin{aligned}
        \text{SLIC} = \sqrt{\frac{\Delta{x}_{ij}^2 + \Delta{y}_{ij}^2}{e^2} + \frac{||y_i - y_j||}{m^2}}
    \end{aligned}
\end{align}
where \(\Delta{x}_{ij}^2 + \Delta{y}_{ij}^2\) is the squared Euclidean distance between two pixels, \(e\) is the search size of SLIC, and \(m\) is a hyperparameter that balances the impact of the pixel distances and
spectra similarity. For each split region, we perform VCA to extract
\(K\) signatures. The resulting endmembers are merged to construct a spectral library with total \(P = KL\), \(L\) is the number of superpixels. In order to get the categorical ID of the total \(KL\) spectral library, a simple but efficient method, K-Means, is used to get the clustering ID. Finally, the spectral library and corresponding ID serve as the training pairs of our proposed model to learn the conditional prior spectral distribution. \autoref{spectral library} shows the construction of the superpixel-based spectral library.

\subsection{Conditional Diffusion model} 

\paragraph{Background of Score-based Diffusion Model.}

In a recent work \cite{song2021scorebased}, score-based \cite{song2019generative} and diffusion-based \cite{ho2020denoising} generative models have been unified into a single continuous-time score-based framework with diffusion driven by stochastic differential equations for the data noising process (i.e., forward SDE) \(\boldsymbol{x}(t)\), \(t \in [0,T]\), \(\boldsymbol{x}(t) \in \mathbb{R}^d \forall t\) in the following form:
\begin{align}
    \begin{aligned}
        d\boldsymbol{x} = -\frac{\beta_t}{2}\boldsymbol{x}dt + \sqrt{\beta(t)}d\boldsymbol{w}
    \end{aligned}
\end{align}
where \(\beta(t): \mathbb{R} \rightarrow \mathbb{R} > 0\) is the noise schedule of the process, and \(\boldsymbol{w}\) is the standard \(d-\)dimension Wiener process. The data distribution is defined when \(t=0\), i.e., \(\boldsymbol{x}(0) \sim p_{\text{data}}\), and a simple, tractable distribution, isotropic Gaussian, is achieved when \(t=T\), i.e., \(\boldsymbol{x}_T \sim \mathcal{N}(\boldsymbol{0}, \boldsymbol{I})\). By performing the following reverse SDE can recover the data-generating distribution:
\begin{align}
    \begin{aligned}
        d\boldsymbol{x} = [-\frac{\beta(t)}{2}\boldsymbol{x} - \beta(t)\nabla_{\boldsymbol{x}_t} \log p_t(\boldsymbol{x}_t)]dt + \sqrt{\beta(t)} d\bar{\boldsymbol{w}}
    \end{aligned}
    \label{eq5}
\end{align}
where \(\bar{\boldsymbol{w}}\) is a standard Wiener process in reverse process. The drift function now depends on the time-dependent score function \(\nabla_{\boldsymbol{x}_t} \log p_t(\boldsymbol{x}_t)\) which is approximated by a neural network \(\boldsymbol{s}_{\theta}\) trained with denoising score matching:
\begin{align}
    \begin{aligned}
        \theta ^{*} = \underset{\theta}{\text{argmin}} \mathbb{E}_{t \sim U(0,1), \boldsymbol{x}_t\sim p(\boldsymbol{x}_t|\boldsymbol{x}_0) \sim p_{\text{data}}} \\\left[\lvert\lvert \boldsymbol{s}_{\theta}(\boldsymbol{x}_t,t) - \nabla_{\boldsymbol{x}_t} \log p(\boldsymbol{x}_t| \boldsymbol{x}_0) \rvert \rvert_2^2\right]
    \end{aligned}
    \label{eq4}
\end{align}
then the approximation \(\nabla_{\boldsymbol{x}_t} \log p_t(\boldsymbol{x}_t) \simeq \boldsymbol{s}_{\theta^*}(\boldsymbol{x}_t, t)\) serves as a plug-in estimate to replace the score function in \autoref{eq5}. Discretization of \autoref{eq5} and solving using Euler-Maruyama discretization amounts to sampling
from the data distribution \(p(\boldsymbol{x})\), the goal of generative modeling.

The above continuous score-matching framework can be extended to conditional generation, as shown in \cite{song2021scorebased}. Suppose we are interested in \(p(\boldsymbol{x}|c)\), where \(c\) is the condition, which can be anything. By using forward diffusion process, \autoref{eq4}, to obtain a family of diffused distributions \(p(\boldsymbol{x}_t|c)\) and the derive the condistional reverse-time SDE:
\begin{align}
    \begin{aligned}
        d\boldsymbol{x} = [-\frac{\beta(t)}{2}\boldsymbol{x} - \beta(t)\nabla_{\boldsymbol{x}_t} \log p_t(\boldsymbol{x}_t|c)]dt + \sqrt{\beta(t)} d\bar{\boldsymbol{w}}
    \end{aligned}
    \label{eq7}
\end{align}
So, \(\nabla_{\boldsymbol{x}_t} \log p_t(\boldsymbol{x}_t|c)\) is all we need to learn in order to be able to sample from \(p(\boldsymbol{x}|c)\) using reverse-time diffusion.
\begin{figure}
    \centering
    \includegraphics[width=\linewidth]{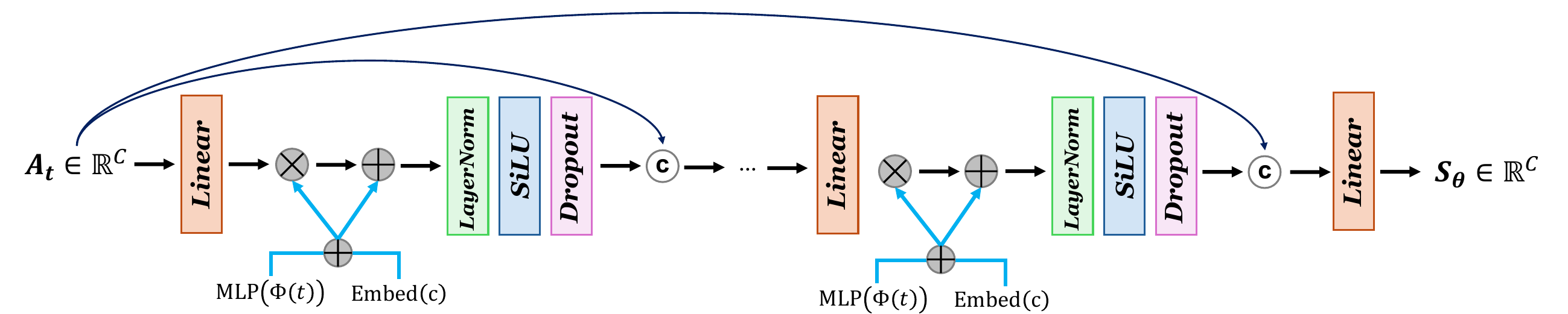}
    \caption{Architecture overview of the MLP-based denoising network \(\boldsymbol{s}_{\theta}(\boldsymbol{A}_t, t, c)\). \(\Phi\) is the Sinusodial Timestep Embedding and Embed(c) represents the label embedding layer.}
    \label{model_arch}
\end{figure}

\paragraph{Forward Process of DPS4Un.} Aiming to learn the spectral prior distribution with guidance, we apply a conditional diffusion model, which has the following forward SDE:
\begin{align}
    \begin{aligned}
        d\boldsymbol{A}_t = -\frac{\beta_t}{2}\boldsymbol{A}_tdt + \sqrt{\beta(t)}d\boldsymbol{w}
    \end{aligned}
\end{align}
where \(\boldsymbol{A}_t = \sqrt{\bar{\alpha}_t}\boldsymbol{A}_0 + \sqrt{1-\bar{\alpha}_t} \boldsymbol{\epsilon}\), and \(\sqrt{\bar{\alpha}_t} = \exp(- \int (\beta_{\tau}/2))d \tau\), \(\boldsymbol{\epsilon} \sim \mathcal{N}(\boldsymbol{0}, \boldsymbol{I})\). The training objective function as follows:
\begin{align}
    \begin{aligned}
        \mathcal{L}(\theta) = \mathbb{E}_{t, \boldsymbol{A}_0, \boldsymbol{\epsilon}, c}\lvert \lvert \boldsymbol{\epsilon} - \sqrt{1-\bar{\alpha}_t} \boldsymbol{s}_{\theta}(\boldsymbol{A}_t, t, c) \rvert \rvert_2^2
    \end{aligned}
\end{align}
Guided by \(\nabla_{\boldsymbol{A}_t} \log p_t(\boldsymbol{A}_t|c)\), we can yield samples from the original spectral prior distribution.

Instead of using UNet or Transformer-like backbone for denoising, we propose a MLP-based denoising network, as shown in \autoref{model_arch}. The denoising networks consist total \(L\) stages followed by a \textit{Linear} layer. The output of each stage concatenates it with the input via a skip connection. Additionally, in order to embed the time information, the denoising network is conditioned on the time step, which performs sinusoidal timestep embedding \(\Phi\) implemented by the MLP layer. The categorical ID of input noisy spectrum is also included, which is implemented by a \textit{labelEmbeder} that converts the discrete condition \(c\) into a continuous embedding vector. Overall, the forward model of the MLP-based denoising network can be formulated as follows:
\begin{align}
    \begin{aligned}
        &\alpha, \gamma = \text{Chunk}(\text{MLP}(t_{emb}, c_{emb})) \\
        &f = \text{Cat}[\boldsymbol{A}_t, \text{Drop}(\text{SiLU}(\text{Linear}(f*(1+\alpha) + \gamma)))]\\
        & \boldsymbol{s}_{\theta}(\boldsymbol{A}_t,t,c) = \text{Linear}(f)
    \end{aligned}
\end{align}
\begin{figure}[!t]
    \centering
    \includegraphics[width=\linewidth]{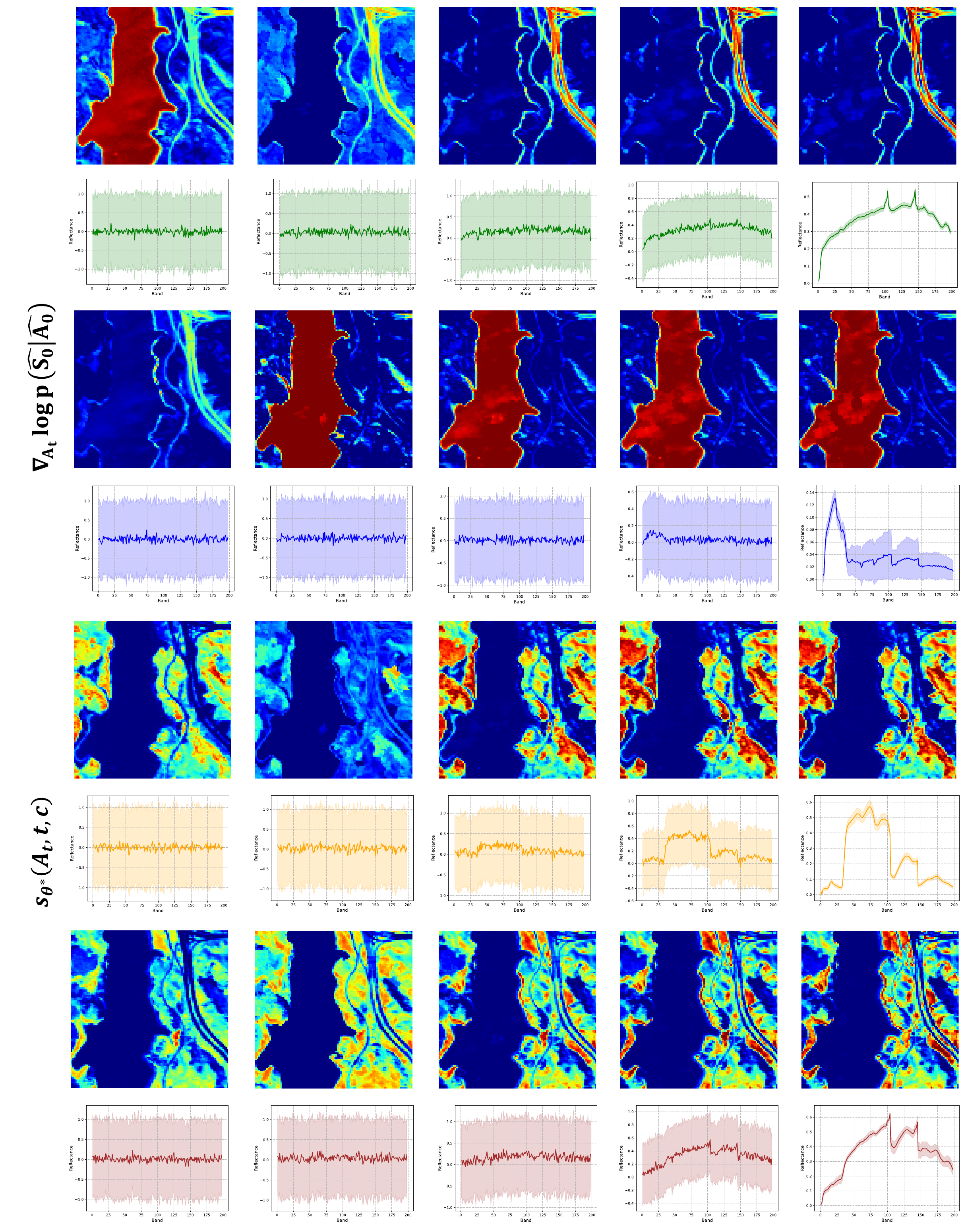}
    \caption{Sampling process (from left to right) of \(\nabla_{\boldsymbol{A}_t} \log p(\hat{\boldsymbol{S}}_0|\hat{\boldsymbol{A}_0})\) (abundance updating, from FCLSU to maximum posterior using DPS4Un), and \(\boldsymbol{s}_{\theta^*} (\boldsymbol{A}_t, t, c)\) (endmember updating, from Gaussian noise to data distribution). The abundance map is initialized from the FCLUS algorithm; basically, it is random. With our pretrained diffusion model under the cluster ID condition, it becomes more fine-grained. Take Jasper Ridge dataset as an example. Each column from time step 951, 701, 451, 201, 1 using DDIM.}
    \label{tmp_gradient}
\end{figure}
\paragraph{Reverse Process for DPS4Un.}
Given that there are two parameters that need to be estimated. Under the Bayesian framework, leveraging the diffusion model as the prior, we can modify \autoref{eq7} to arrive at the reverse diffusion sampler for sampling from the posterior distribution:
\begin{align}
    \begin{aligned}
        d\boldsymbol{A_t} = [-\frac{\beta(t)}{2}\boldsymbol{A}_t - \beta(t)(\nabla_{\boldsymbol{A}_t} \log p_t(\boldsymbol{X}|\boldsymbol{A}_t, \boldsymbol{S}_t) + \\ 
        \nabla_{\boldsymbol{A}_t} \log p_t(\boldsymbol{S}_t|\boldsymbol{A}_t) + \nabla \log p_t (\boldsymbol{A}_t|c)]dt + \sqrt{\beta(t)} d\bar{\boldsymbol{w}}
    \end{aligned}
    \label{eq11}
\end{align}
where we have used the fact that:
\begin{align}
    \begin{aligned}
        \nabla_{\boldsymbol{A}_t} \log p_t (\boldsymbol{A}_t|\boldsymbol{X}, \boldsymbol{S}_t,c) = \nabla_{\boldsymbol{A}_t} \log p_t(\boldsymbol{X}|\boldsymbol{A}_t, \boldsymbol{S}_t) \\
        + \nabla_{\boldsymbol{A}_t} \log p_t(\boldsymbol{S}_t|\boldsymbol{A}_t) + \nabla \log p_t (\boldsymbol{A}_t|c)
    \end{aligned}
    \label{eq12}
\end{align}
In \autoref{eq11}, three terms need to be calculated: the conditional score function \(\nabla_{\boldsymbol{A}_t} \log p_t (\boldsymbol{A}_t|c)\), the abundance estimation \(\nabla_{\boldsymbol{A}_t} \log p_t(\boldsymbol{S}_t|\boldsymbol{A}_t)\), and the likelihood \(\nabla_{\boldsymbol{A}_t} \log p_t(\boldsymbol{X}|\boldsymbol{A}_t, \boldsymbol{S}_t)\). To compute the former term involving \(p_t(\boldsymbol{A}|c)\), we can simply use the pre-trained condition score-matching function \(\boldsymbol{s}_{\theta^*} (\boldsymbol{A}_t, t, c)\). However, the latter two terms is hard to acquire in closed-form due to the dependence of the time \(t\). In order to solve this problem, consider the following in the forward diffusion:
\begin{align}
    \begin{aligned}
        \boldsymbol{A}_t = \sqrt{\bar{\alpha}_t}\boldsymbol{A}_0 + \sqrt{1-\bar{\alpha}_t} \boldsymbol{\epsilon}, \boldsymbol{\epsilon} \sim \mathcal{N}(\boldsymbol{0}, \boldsymbol{I})
    \end{aligned}
\end{align}
we can obtain the specialized representation of the posterior mean of \(\boldsymbol{A}_0\) at \(t\) given \(\boldsymbol{A}_t\) through Tweedie’s approach \cite{efron2011tweedie, kim2021noise2score}, derived as follows:
\begin{align}
    \begin{aligned}
        \hat{\boldsymbol{A}}_0 = \mathbb{E}[\boldsymbol{A}_0 | \boldsymbol{A}_t]&=\frac{1}{\sqrt{\bar{\alpha}_t}}(\boldsymbol{A}_t + (1-\bar{\alpha}_t) \nabla_{\boldsymbol{A}_t} \log p_t(\boldsymbol{A}_t|c))\\
        &\simeq\frac{1}{\sqrt{\bar{\alpha}_t}}(\boldsymbol{A}_t + (1-\bar{\alpha}_t) \boldsymbol{s}_{\theta}(\boldsymbol{A}_t,t,c))
    \end{aligned}
\end{align}
Given the posterior mean \(\hat{\boldsymbol{A}}_0\) that can be efficiently computed at the intermediate steps, the next step is to provide a tractable approximation for abundance estimation and likelihood. For the abundance estimation term, \(\nabla_{\boldsymbol{A}_t} \log p_t(\boldsymbol{S}_t|\boldsymbol{A}_t)\), since we have a optimal estimation of ground truth endmember matrix \(\hat{\boldsymbol{A}}_0\), the optimal abudance matrix will be deterministic given the endmember. Hence, \(p(\boldsymbol{S}_t|\boldsymbol{A}_t)\) can be approximated by \(p(\hat{\boldsymbol{S}_0} | \hat{\boldsymbol{A}_0})\). Given the above approximation, the data likelihood with the following approximation:
\begin{align}
    \begin{aligned}
        p(\boldsymbol{X}|\boldsymbol{A}_t, \boldsymbol{S}_t) \simeq p(\boldsymbol{X}|\hat{\boldsymbol{A}}_0, \hat{\boldsymbol{S}}_0) \propto \exp(-\frac{||\boldsymbol{X} - \hat{\boldsymbol{A}}_0 \hat{\boldsymbol{S}_0}||_2^2}{2 \sigma_c^2})
    \end{aligned} 
\end{align}
Then the approximate gradient of the log likelihood is:
\begin{align}
    \begin{aligned}
        \nabla_{\boldsymbol{A}_t} \log p(\boldsymbol{X}|\boldsymbol{A}_t, \boldsymbol{S}_t) \simeq \nabla_{\boldsymbol{A}_t} \log p(\boldsymbol{X}|\hat{\boldsymbol{A}_0}, \hat{\boldsymbol{S}_0})
    \end{aligned}
    \label{eq16}
\end{align}
The approximation gradient of the log abundance estimation:
\begin{align}
    \begin{aligned}
        \nabla_{\boldsymbol{A}_t} \log p(\boldsymbol{S}_t|\boldsymbol{A}_t) \simeq \nabla_{\boldsymbol{A}_t} \log p(\hat{\boldsymbol{S}_0}|\hat{\boldsymbol{A}_0})
    \end{aligned}
    \label{eq17}
\end{align}
Now, these two terms are analytically tractable.

By differentiating \(p(\boldsymbol{X}|\boldsymbol{A}_t, \boldsymbol{S}_t)\) with respect to \(\boldsymbol{A}_t\) using \autoref{eq16}, we can get:
\begin{align}
    \begin{aligned}
        \nabla_{\boldsymbol{A}_t} \log p(\boldsymbol{X}|\boldsymbol{A}_t, \boldsymbol{S}_t) \simeq -\frac{1}{\sigma_c^2} \nabla_{\boldsymbol{A}_t} ||\boldsymbol{X} - \hat{\boldsymbol{A}_0}(\boldsymbol{A}_t) \hat{\boldsymbol{S}_0}||_2^2
    \end{aligned}
    \label{eq18}
\end{align}
Consequently, plug in the results from \autoref{eq17} and \autoref{eq18} to \autoref{eq12} with the trained condition score function, we can finally conclude that:
\begin{align}
    \begin{aligned}
        \nabla_{\boldsymbol{A}_t} \log p_t (\boldsymbol{A}_t|\boldsymbol{X}, \boldsymbol{S}_t,c)  \simeq \boldsymbol{s}_{\theta^*}(\boldsymbol{A}_t, t, c) \\
        - \frac{1}{\sigma_c^2}\nabla_{\boldsymbol{A}_t} ||\boldsymbol{X} - \hat{\boldsymbol{A}_0} \hat{\boldsymbol{S}_0}||_2^2 - \nabla_{\boldsymbol{A}_t} \log p(\hat{\boldsymbol{S}_0}|\hat{\boldsymbol{A}_0})
    \end{aligned}
    \label{eq19}
\end{align}
For a Linear system, considering the rank of endmember \(\boldsymbol{A} \in \mathbb{R}^{C \times K}\) is equal to \(K\) (the number of endmember \(K\) is usually much smaller than number of bands \(C\), \(K \ll C\)), meaning that \(\boldsymbol{A}\) is a full row matrix, the least squares solution can be described as follows:
\begin{align}
    \begin{aligned}
        \hat{\boldsymbol{S}} = \boldsymbol{A}^T (\boldsymbol{A} \boldsymbol{A}^T)^{-1} \boldsymbol{X}
    \end{aligned}
\end{align}
In practice, we choose projected gradient descent methods to update the abundance \(\hat{\boldsymbol{S}}_0\) given \(\hat{\boldsymbol{A}}_0\) and \(\boldsymbol{X}\) to solve \(\nabla_{\boldsymbol{A}_t} \log p(\hat{\boldsymbol{S}_0}|\hat{\boldsymbol{A}_0})\), as follows:
\begin{align}
    \begin{aligned}
        \hat{\boldsymbol{S}_0} = \hat{\boldsymbol{S}_0} - \lambda (\hat{\boldsymbol{A}_0} \hat{\boldsymbol{A}}_0^T \hat{\boldsymbol{S}_0} - \hat{\boldsymbol{A}_0} \boldsymbol{X})
    \end{aligned}
\end{align}
\(\lambda\) is the step of gradient descent. Overall, the sapling strategy is summarized in \autoref{algo}, 
We use DDIM \cite{song2021DDIM} to accelerate the sampling speed, as shown in \autoref{eq22}. It should be noted that \(\boldsymbol{X}\) has been split into superpixel level with total \(L\) sub-regions, and all the gradient steps in \autoref{eq19} are computed within each region-based area.  \autoref{tmp_gradient} visualizes the inference process of DPS4Un, which clearly shows the updating progress of endmember and abundance. Even though the abundance initialization is random using FCLSU \cite{heylen2011fully}, the gradient still can flow to the desired distribution since our model conditions on the categorical label, reducing the randomness of generation of the unconditional diffusion model. 
\begin{align}
    \begin{aligned}
    \boldsymbol{A}_{i-1}^{'} = \sqrt{\alpha_{i-1}} (\frac{\boldsymbol{A}_{i} - \sqrt{1-\alpha_i} \hat{\boldsymbol{s}}_{\theta}}{\sqrt{\alpha_i}})
                    \\
                    + \sqrt{1 - \alpha_{i-1} - \sigma_{i}^{2}}                         \hat{\boldsymbol{s}}_{\theta} + \tilde{\sigma}_{i}                            \boldsymbol{z},  \boldsymbol{z} \sim \mathcal{N}(\boldsymbol{0}, \boldsymbol{I})
    \end{aligned}
    \label{eq22}
\end{align}

\begin{algorithm}
\caption{Posterior Sampling of DPS4Un with Spectral Variability Consideration}
    \begin{algorithmic}[1] \label{algo}
        \REQUIRE DDIM time step \(T\), \(\boldsymbol{X}\), initialized abundance \(\hat{\boldsymbol{S}}_0\), \(\{\tilde{\sigma}_i{}\}_{i=1}^{T}\), condition \(c\), step of gradient decent \(\lambda\), Superpixel region set \(\{R_l\}_{l=n}^{L}\)
        \STATE \(\boldsymbol{A}_T \sim \mathcal{N}(\boldsymbol{0}, \boldsymbol{I}) \in \mathbb{R}^{L \times K \times C}\)
         \FOR{$i = T, ..., 1$}
         \STATE \(\hat{\boldsymbol{s}} \leftarrow \boldsymbol{s}(\boldsymbol{A}_i,i, c)\)
         \STATE \(\hat{\boldsymbol{A}_0} \leftarrow \frac{1}{\sqrt{\bar{\alpha_i}}} (\boldsymbol{A}_i + (1-\bar{\alpha}_i) \hat{\boldsymbol{s}})\)
        \STATE \(\boldsymbol{z} \sim \mathcal{N}(\boldsymbol{0}, \boldsymbol{I}) \in \mathbb{R}^{L \times K \times C}\)
        \STATE \(\boldsymbol{A}_{i-1}^{'} = \text{DDIM}(\boldsymbol{A}_i, \alpha_i, \hat{\boldsymbol{s}}_{\theta}, \tilde{\sigma}_i, \boldsymbol{z})\)
            \FOR{$l = L, .., 1$}
                \STATE \(\hat{\boldsymbol{S}}_0^{l} = \hat{\boldsymbol{S}_0^l} - \lambda (\hat{\boldsymbol{A}}_0^l \hat{\boldsymbol{A}}_0^l{^T} \hat{\boldsymbol{S}}_0^l - \hat{\boldsymbol{A}}_0^l \boldsymbol{X}^l)\)
                \STATE \({\boldsymbol{A}_{i-1}^l}^{'} = {\boldsymbol{A}_{i-1}^l}^{'} - \nabla_{\boldsymbol{A}_t}||\boldsymbol{X}^l - \hat{\boldsymbol{A}}_0^l \hat{\boldsymbol{S}}_0^l||_2^2\)
            \ENDFOR
        \ENDFOR
        \RETURN \(\hat{\boldsymbol{A}_0}\) and \(\hat{\boldsymbol{S}_0}\)
    \end{algorithmic}
\end{algorithm}

%% file: 4_Results_and_Analysis.tex
\section{Results and Analysis}
\label{sec:results}

\begin{figure*}[!t]
    \centering
    \includegraphics[width=\textwidth]{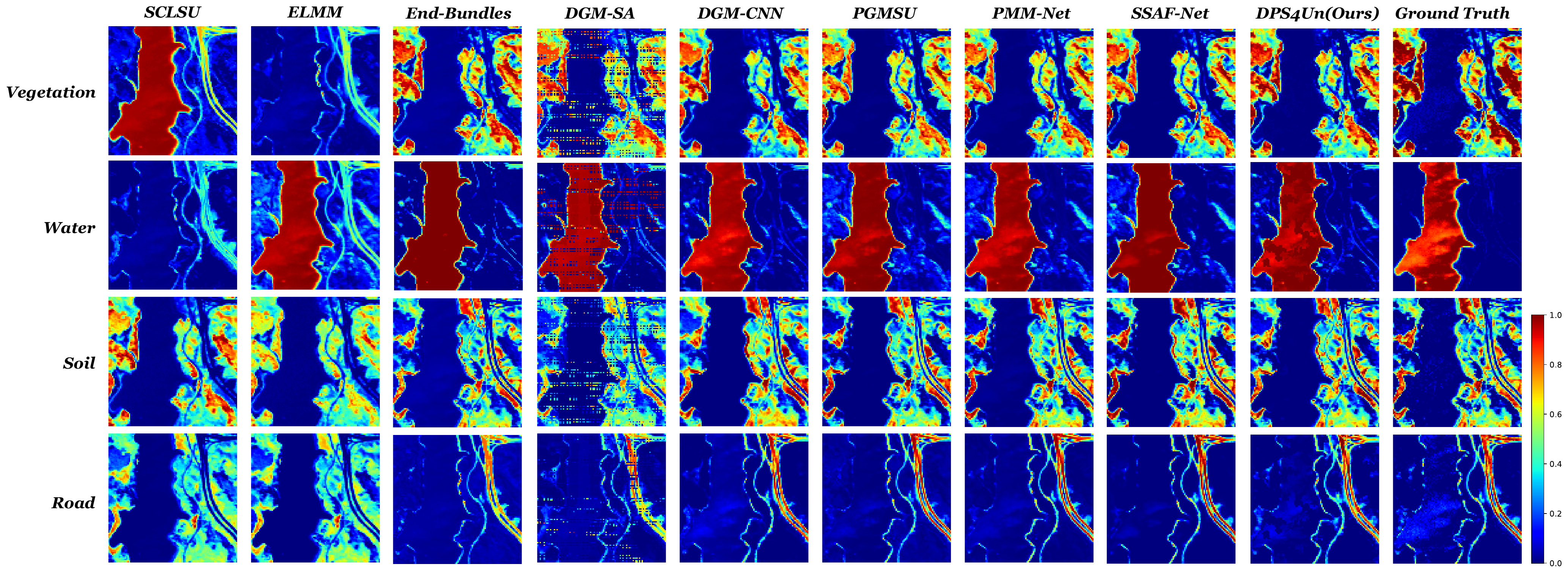}
    \caption{Estimated abundances on the Jasper Ridge dataset.}
    \label{jasper_figures}
\end{figure*}

\begin{figure*}[!h]
    \centering
    \includegraphics[width=\textwidth]{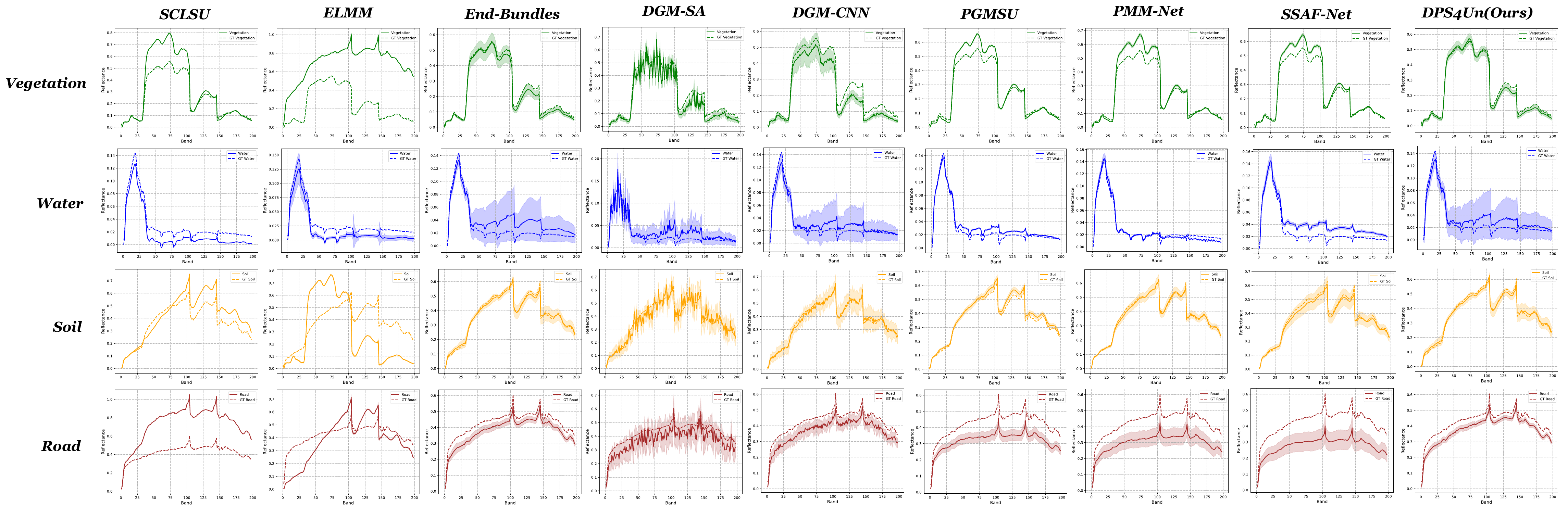}
    \caption{Estimated endmembers on the Jasper Ridge dataset. Dash line: ground truth, solid line (or with \(\pm \sigma\) shadow area): estimated endmembers.}
    \label{jasper_curevs}
\end{figure*}

\subsection{Implementation Details and Datasets}

\paragraph{Implementation Details.}
The proposed DPS4Un model undergoes two stages. Pretrain the conditional diffusion prior learner, and fix the pre-trained diffusion model for posterior sampling. All
experiments are conducted using PyTorch 1.10.2 platform on a NVIDIA RTX A6000 Ada Generation with 48GB of VRAM. The epoch of pre-training is set as 100000, and the training time consumption is around 10 minutes, and the sampling time consumption is around 20 seconds on average for all datasets. The DDIM sampling steps are set as 20.

\begin{table*}[!b]
\caption{Experimental Results of Different Methods on the Jasper Ridge Dataset. Top results are highlighted as \colorbox{mycolor}{\textbf{best}}, \colorbox{mycolor2}{second}, and \colorbox{mycolor3}{third}.}
\label{acc_JR}
\resizebox{\textwidth}{!}{
\begin{tabular}{cc|ccccccccc}
\hline
\multicolumn{1}{l}{}     & \multicolumn{1}{l|}{} & \multicolumn{9}{c}{Spectral Variability}                                                                                                \\ \hline
\multirow{2}{*}{Metrics} & \multirow{2}{*}{\#EM} & \Checkmark    & \Checkmark    & \Checkmark                                                          & \Checkmark    & \Checkmark     & \Checkmark    & \Checkmark     & \Checkmark     & \Checkmark    \\
                         &                       & SCLSU  & ELMM   & \begin{tabular}[c]{@{}c@{}}End-Bundles\end{tabular} & DGM-SA & DGM-CNN & PGMSU  & PPM-Net & SSF-Net & Ours   \\ \hline
\multirow{5}{*}{RMSE $\downarrow$}    & Vegetation            & 0.5115 & 0.4914 & \colorbox{mycolor}{\textbf{0.0565}}                                                       & 0.2128 & 0.1137  & 0.1036 & 0.1058  & \colorbox{mycolor3}{0.0925}  & \colorbox{mycolor2}{0.0627} \\
                         & Water                 & 0.6548 & 0.6412 & 0.1063                                                       & 0.2603 & \colorbox{mycolor}{\textbf{0.0881}}  & 0.1204 & \colorbox{mycolor3}{0.1061}  & 0.1165  & \colorbox{mycolor2}{0.0956} \\
                         & Soil                  & 0.3558 & 0.3669 & \colorbox{mycolor2}{0.0804}                                                       & 0.1798 & 0.1308  & \colorbox{mycolor3}{0.0805} & 0.0851  & 0.0821  & \colorbox{mycolor}{\textbf{0.0727}} \\
                         & Road                  & 0.2437 & 0.2489 & 0.0657                                                       & 0.1768 & 0.0659  & \colorbox{mycolor2}{0.0583} & \colorbox{mycolor}{\textbf{0.0548}}  & 0.0693  & \colorbox{mycolor3}{0.0622} \\
                         & aRMSE                 & 0.4181 & 0.4134 & \colorbox{mycolor2}{0.0567}                                                       & 0.1267 & 0.0797  & 0.0691 & 0.0670   & \colorbox{mycolor3}{0.0647}  & \colorbox{mycolor}{\textbf{0.0517}} \\ \hline
\multirow{5}{*}{SAD $\downarrow$}     & Vegetation            & 0.1481 & 0.1481 & \colorbox{mycolor2}{0.0535}                                                       & 0.1816 & 0.1075  & \colorbox{mycolor3}{0.0584} & 0.0628  & \colorbox{mycolor}{\textbf{0.0413}}  & 0.0592 \\
                         & Water                 & 0.2582 & 0.2565 & 0.2916                                                       & 0.3340  & 0.1917  & \colorbox{mycolor}{\textbf{0.1366}} & \colorbox{mycolor3}{0.1764}  & 0.2425  & \colorbox{mycolor2}{0.1663} \\
                         & Soil                  & 0.1166 & 0.1171 & \colorbox{mycolor3}{0.0260}                                                        & 0.1052 & 0.0400    & 0.0298 & \colorbox{mycolor2}{0.0225}  & \colorbox{mycolor}{\textbf{0.0213}}  & 0.0423 \\
                         & Road                  & 0.0901 & 0.0828 & 0.0426                                                       & 0.0862 & 0.0434  & \colorbox{mycolor3}{0.0290}  & \colorbox{mycolor2}{0.0256}  & \colorbox{mycolor}{\textbf{0.0246}}  & 0.0464 \\
                         & aSAD                  & 0.1533 & 0.1511 & 0.1034                                                       & 0.1768 & 0.0957  & \colorbox{mycolor}{\textbf{0.0634}} & \colorbox{mycolor2}{0.0718}  & 0.0824  & \colorbox{mycolor3}{0.0785} \\ \hline
\end{tabular}
}
\end{table*}

\paragraph{Datasets} Three widely used real-world datasets are selected for comparison.

\begin{itemize}
    \item \textit{Jasper Ridge Dataset} This dataset \cite{zhu2017hyperspectralunmixinggroundtruth} is popular in the unmixing task. Each pixel is recorded at 224 channels, ranging from 380 nm to 2500 nm. Usually, pre-processed data are adopted by selecting ROI and 26 channels have been removed due to the water absorption and atmospheric effects. Finally, a region with 100 by 100 pixels with 198 channels is chosen. Four endmembers in this dataset. Total 228 superpixels are split using SLIC.

    \item \textit{Urban Dataset} For this data \cite{zhu2017hyperspectralunmixinggroundtruth}, there are 210 wavelengths ranging from 400 nm to 2500 nm, resulting in a spectral resolution of 10 nm. To obtain data with an acceptable SNR, and 47 channels has been removed. The data we used contains 307 by 307 pixels with 162 bands. Six endmembers in this dataset. Total 558 superpixels are split using SLIC.

    \item \textit{SMScene Dataset} It is a new and highly complex designed scene benchmark \cite{10016626} where high-resolution images of was used to obtain the abundance ground truth, with resolution 0.125 mm/pixel. The mixed hyperspectral image, has a spatial resolution of 16 × 0.125 mm/pixel, with a 64 \(\times\) 128 \(\times\) 132 image size. Point, line, and polygon features are the basic elements of this dataset, including stones, sticks, leaves, and moss. There are four endmembers in this scene. Total 226 superpixels are split using SLIC.
\end{itemize}
\begin{figure*}[!h]
    \centering
    \includegraphics[width=\textwidth]{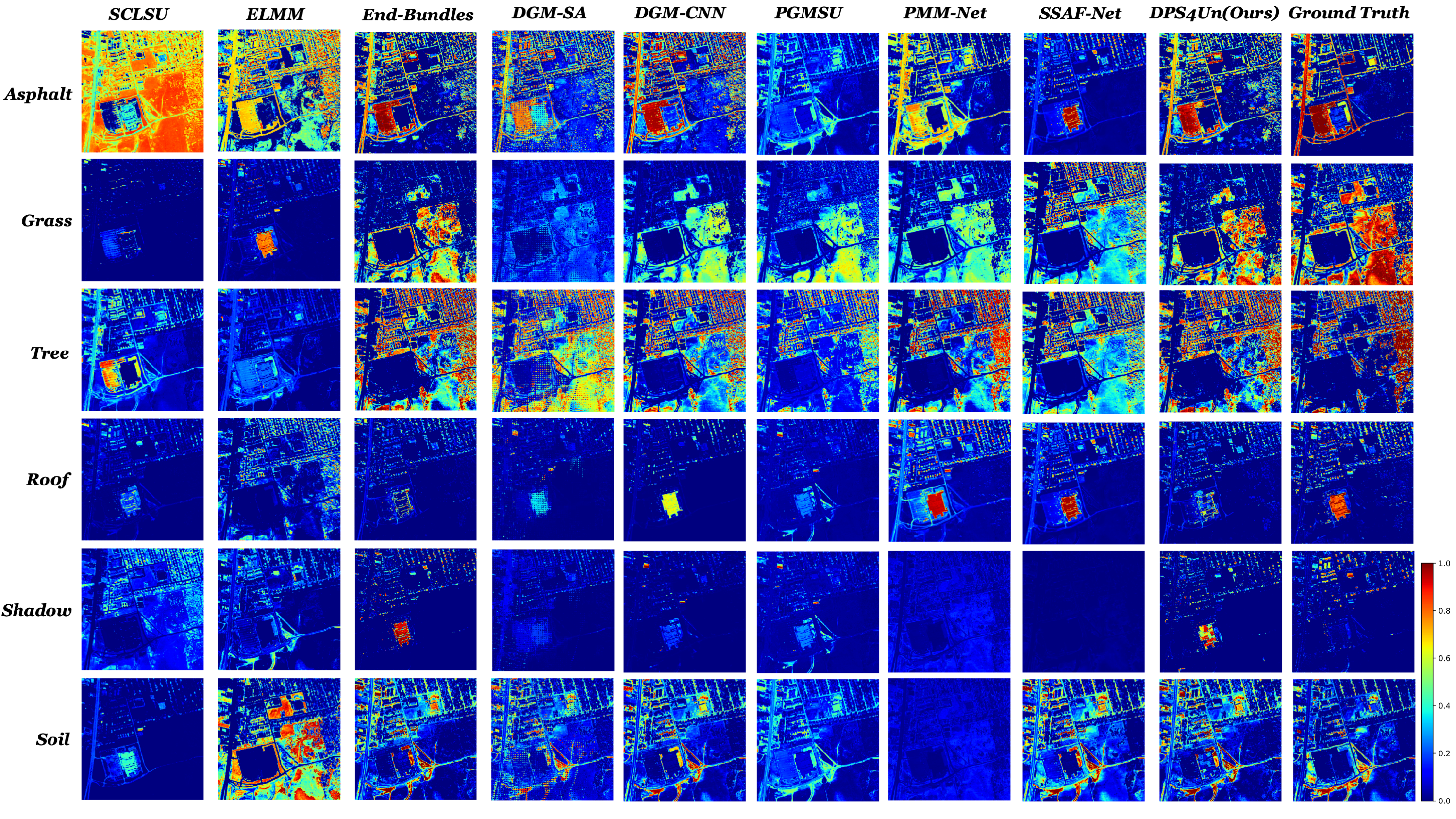}
    \caption{Estimated abundances on the Urban dataset.}
    \label{urban_figures}
\end{figure*}
\begin{figure*}[!h]
    \centering
    \includegraphics[width=\textwidth]{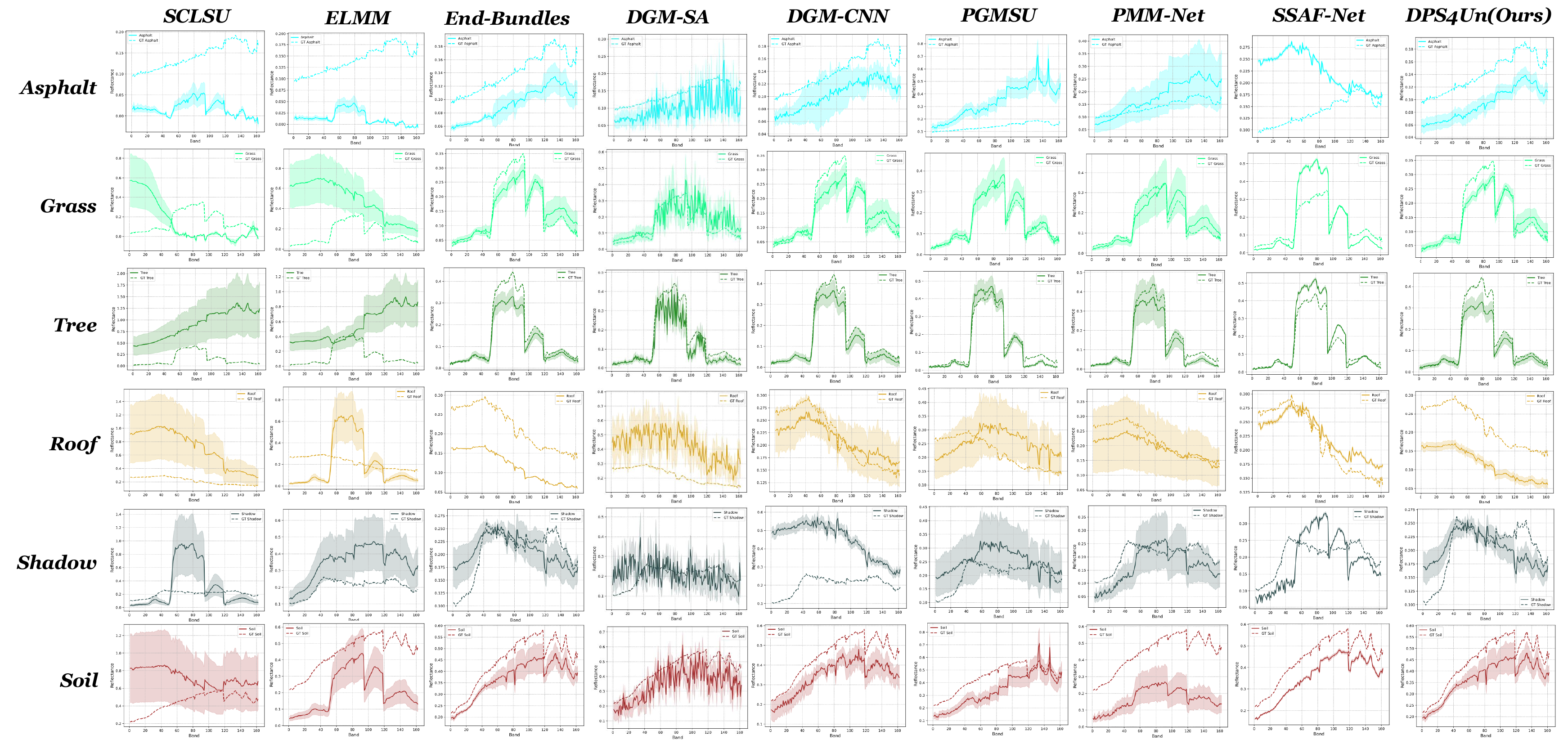}
    \caption{Estimated endmembers on the Urban dataset. Dash line: ground truth, solid line (or with \(\pm \sigma\) shadow area): estimated endmembers.}
    \label{urban_curves}
\end{figure*}
\begin{table*}[!t]
\caption{Experimental Results of Different Methods on the Urban Dataset. Top results are highlighted as \colorbox{mycolor}{\textbf{best}}, \colorbox{mycolor2}{second}, and \colorbox{mycolor3}{third}.}
\resizebox{\textwidth}{!}{
\begin{tabular}{cc|ccccccccc}
\hline
\multicolumn{1}{l}{}     & \multicolumn{1}{l|}{} & \multicolumn{9}{c}{Spectral Variability}                                                                                              \\ \hline
\multirow{2}{*}{Metrics} & \multirow{2}{*}{\#EM} & \Checkmark    & \Checkmark    & \Checkmark                                                        & \Checkmark    & \Checkmark     & \Checkmark    & \Checkmark     & \Checkmark     & \Checkmark    \\
                         &                       & SCLSU  & ELMM   & \begin{tabular}[c]{@{}c@{}}End-Bundles\end{tabular} & DGM-SA & DGM-CNN & PGMSU  & PPM-Net & SSF-Net & Ours   \\ \hline
\multirow{7}{*}{RMSE $\downarrow$}    & Asphalt               & \colorbox{mycolor}{\textbf{0.2159}} & 0.2797 & 0.2508                                                     & 0.3062 & 0.3141  & 0.3218 & \colorbox{mycolor2}{0.2204}  & 0.3292  & \colorbox{mycolor3}{0.2461} \\
                         & Grass                 & 0.4167 & \colorbox{mycolor}{\textbf{0.1852}} & 0.2855                                                     & 0.3806 & \colorbox{mycolor3}{0.2687}  & 0.5239 & \colorbox{mycolor2}{0.2384}  & 0.3724  & 0.2741 \\
                         & Tree                  & 0.2779 & 0.2754 & 0.2580                                                     & 0.3342 & \colorbox{mycolor2}{0.2389}  & 0.2658 & \colorbox{mycolor}{\textbf{0.1475}}  & 0.2844  & \colorbox{mycolor3}{0.2455} \\
                         & Roof                  & 0.1751 & \colorbox{mycolor}{\textbf{0.1099}} & 0.1620                                                     & 0.1824 & 0.1522  & 0.1731 & \colorbox{mycolor2}{0.1325}  & \colorbox{mycolor3}{0.1423}  & 0.1476 \\
                         & Shadow                & 0.1396 & 0.4693 & 0.1755                                                     & \colorbox{mycolor}{\textbf{0.1126}} & 0.1211  & \colorbox{mycolor2}{0.1163} & 0.1264  & \colorbox{mycolor3}{0.1179}  & 0.1474 \\
                         & Soil                  & 0.2548 & 0.1642 & 0.1834                                                     & 0.1663 & \colorbox{mycolor2}{0.1491}  & 0.2818 & 0.2184  & \colorbox{mycolor2}{0.1597}  & \colorbox{mycolor}{\textbf{0.1384}} \\
                         & aRMSE                 & 0.2282 & 0.2373 & 0.1845                                                     & \colorbox{mycolor2}{0.1663} & 0.1853  & 0.2781 & \colorbox{mycolor}{\textbf{0.1616}}  & 0.2405  & \colorbox{mycolor3}{0.1751} \\ \hline
\multirow{7}{*}{SAD $\downarrow$}     & Asphalt               & 0.1339 & 0.2050 & \colorbox{mycolor2}{0.0573}                                                     & 0.2843 & \colorbox{mycolor}{\textbf{0.0515}}  & 0.1395 & 0.1520 & 0.3405  & \colorbox{mycolor3}{0.0731} \\
                         & Grass                 & 0.4134 & \colorbox{mycolor}{\textbf{0.1137}} & 0.1859                                                     & 0.3310 & \colorbox{mycolor3}{0.1813}  & 0.6859 & 0.1992  & 0.7747  & \colorbox{mycolor2}{0.1676} \\
                         & Tree                  & 0.1097 & 0.1097 & \colorbox{mycolor3}{0.0663}                                                     & 0.2237 & \colorbox{mycolor2}{0.0641}  & 0.0819 & 0.0862  & 0.0744  & \colorbox{mycolor}{\textbf{0.0634}} \\
                         & Roof                  & 0.1397 & 0.1397 & \colorbox{mycolor3}{0.0943}                                                     & 0.1725 & 0.1090  & 0.0962 & \colorbox{mycolor}{\textbf{0.0904}}  & 0.0957  & \colorbox{mycolor2}{0.0941} \\
                         & Shadow                & 0.2885 & 0.8104 & \colorbox{mycolor3}{0.1788}                                                     & 0.3436 & 0.3081  & \colorbox{mycolor}{\textbf{0.1115}} & 0.2598  & 0.3103  & \colorbox{mycolor2}{0.1744} \\
                         & Soil                  & 1.1953 & 0.0928 & \colorbox{mycolor2}{0.0292}                                                     & 0.1666 & \colorbox{mycolor3}{0.0465}  & 0.2121 & 0.2164  & 0.0471  & \colorbox{mycolor}{\textbf{0.0274}} \\
                         & aSAD                  & 0.3801 & 0.2452 & \colorbox{mycolor2}{0.1020}                                                     & 0.2536 & \colorbox{mycolor3}{0.1267}  & 0.2212 & 0.1673  & 0.1906  & \colorbox{mycolor}{\textbf{0.0997}} \\ \hline
\end{tabular}
}
\label{acc_urban}
\end{table*}

\begin{table*}[]
\caption{Experimental Results of Different Methods on the SMScene Dataset. Top results are highlighted as \colorbox{mycolor}{\textbf{best}}, \colorbox{mycolor2}{second}, and \colorbox{mycolor3}{third}.}
\resizebox{\textwidth}{!}{
\begin{tabular}{cc|ccccccccc}
\hline
\multicolumn{1}{l}{}     & \multicolumn{1}{l|}{} & \multicolumn{9}{c}{Spectral   Variability}                                                                                             \\ \hline
\multirow{2}{*}{Metrics} & \multirow{2}{*}{\#EM} & \Checkmark    & \Checkmark    & \Checkmark                                                         & \Checkmark    & \Checkmark     & \Checkmark    & \Checkmark     & \Checkmark     & \Checkmark    \\
                         &                       & SCLSU  & ELMM   & \begin{tabular}[c]{@{}c@{}}End-Bundles\end{tabular} & DGM-SA & DGM-CNN & PGMSU  & PPM-Net & SSF-Net & Ours   \\ \hline
\multirow{5}{*}{RMSE $\downarrow$}    & Moss                  & 0.6182 & 0.5563 & \colorbox{mycolor3}{0.4835}                                                      & 0.8287 & \colorbox{mycolor2}{0.4697}  & 0.5864 & 0.8703  & 0.6173  & \colorbox{mycolor}{\textbf{0.4105}} \\
                         & Pebbles               & \colorbox{mycolor}{\textbf{0.0484}} & \colorbox{mycolor2}{0.0526} & 0.1598                                                      & 0.6223 & 0.1360  & 0.1361 & 0.4258  & 0.1042  & \colorbox{mycolor3}{0.0542} \\
                         & Sticks                & 0.3391 & 0.2074 & 0.2154                                                      & \colorbox{mycolor2}{\textbf{0.1394}} & \colorbox{mycolor}{\textbf{0.1240}}  & 0.1677 & 0.1795  & \colorbox{mycolor3}{0.1412}  & 0.1638 \\
                         & Leaves                & 0.2728 & 0.4256 & 0.3537                                                      & 0.2972 & 0.3919  & \colorbox{mycolor2}{0.2204} & \colorbox{mycolor}{\textbf{0.2144}}  & \colorbox{mycolor3}{0.2474}  & 0.2817 \\
                         & aRMSE                 & 0.3614 & 0.3518 & 0.2668                                                      & 0.5358 & \colorbox{mycolor3}{0.2600}  & 0.3025 & 0.4826  & \colorbox{mycolor}{\textbf{0.1815}}  & \colorbox{mycolor2}{0.2098} \\ \hline
\multirow{5}{*}{SAD $\downarrow$}     & Moss                  & 0.1097 & 0.0865 & 0.0791                                                      & 0.1684 & \colorbox{mycolor2}{0.0556}  & 0.0936 & \colorbox{mycolor}{\textbf{0.0549}}  & \colorbox{mycolor3}{0.0770}  & 0.0825 \\
                         & Pebbles               & \colorbox{mycolor2}{0.0475} & \colorbox{mycolor3}{0.0475} & 0.0504                                                      & 0.5969 & 0.4362  & 0.1014 & 0.5721  & 0.2228  & \colorbox{mycolor}{\textbf{0.0403}} \\
                         & Sticks                & 0.4826 & 0.4826 & 0.1588                                                      & 0.1564 & \colorbox{mycolor2}{0.1074}  & 0.1712 & 0.2647  & \colorbox{mycolor}{\textbf{0.0697}}  & \colorbox{mycolor3}{0.1456} \\
                         & Leaves                & \colorbox{mycolor3}{0.0256} & 0.1701 & \colorbox{mycolor2}{0.0126}                                                      & 0.1062 & 0.0329  & 0.0985 & 0.0339  & 0.2817  & \colorbox{mycolor}{\textbf{0.0118}} \\
                         & aSAD                  & 0.1638 & 0.1942 & \colorbox{mycolor2}{0.0750}                                                      & 0.2570 & 0.1580  & \colorbox{mycolor3}{0.1162} & 0.2314  & 0.1627  & \colorbox{mycolor}{\textbf{0.0704}} \\ \hline
\end{tabular}
}
\label{acc_RMM}
\end{table*}

\subsection{Performance Evaluation Criteria}
To quantitatively assess the unmixing performance, we use root mean square error (RMSE) and the overall mean RMSE (aRMSE) of the abundance corresponding to different endmembers. These metrics are defined in \autoref{eq23}. To comprehensively assess the similarity between the proposed model and the true hyperspectral data in terms of endmembers, we introduced the SAD and its overall mean SAD (aSAD) as additional metrics in \autoref{eq24}.
\begin{align}
    \begin{aligned}
        &\text{RMSE}_{\boldsymbol{S}_k} = \sqrt{\frac{1}{N} \underset{n=1}{\overset{N}{\sum}} ||\boldsymbol{s}_n - \hat{\boldsymbol{s}}_n||^2_2} \\
        &\text{aRMSE}_{\boldsymbol{S}} = \frac{1}{K}\underset{k=1}{\overset{K}{\sum}} \text{RMSE}_{\boldsymbol{S}_k}
    \end{aligned}
    \label{eq23}
\end{align}
\begin{align}
    \begin{aligned}
        &\text{SAD}_{\boldsymbol{A}_k} = \frac{1}{N}\underset{n=1}{\overset{N}{\sum}} \arccos \Bigg( \frac{\boldsymbol{A}_n^T \hat{\boldsymbol{A}_k^T}}{||\boldsymbol{A}_n||_2 ||\hat{\boldsymbol{A}_k^T}||_2} \Bigg) \\
        &\text{aSAD} = \frac{1}{K}\underset{k=1}{\overset{K}{\sum}} \text{SAD}_{\boldsymbol{A}_k}
    \end{aligned}
    \label{eq24}
\end{align}
Since our method can model the spectral variability, to benchmark against the reference, the average values of the ensemble endmembers are used. Also, it should be noticed that using the mean endmember value to represent the "pure endmember" to compare with the ground truth endmember may lead to some bias in these metrics. So, we should put more emphasis on the abundance map visually.
\begin{figure}[!h]
    \centering
    \includegraphics[width=0.49\textwidth]{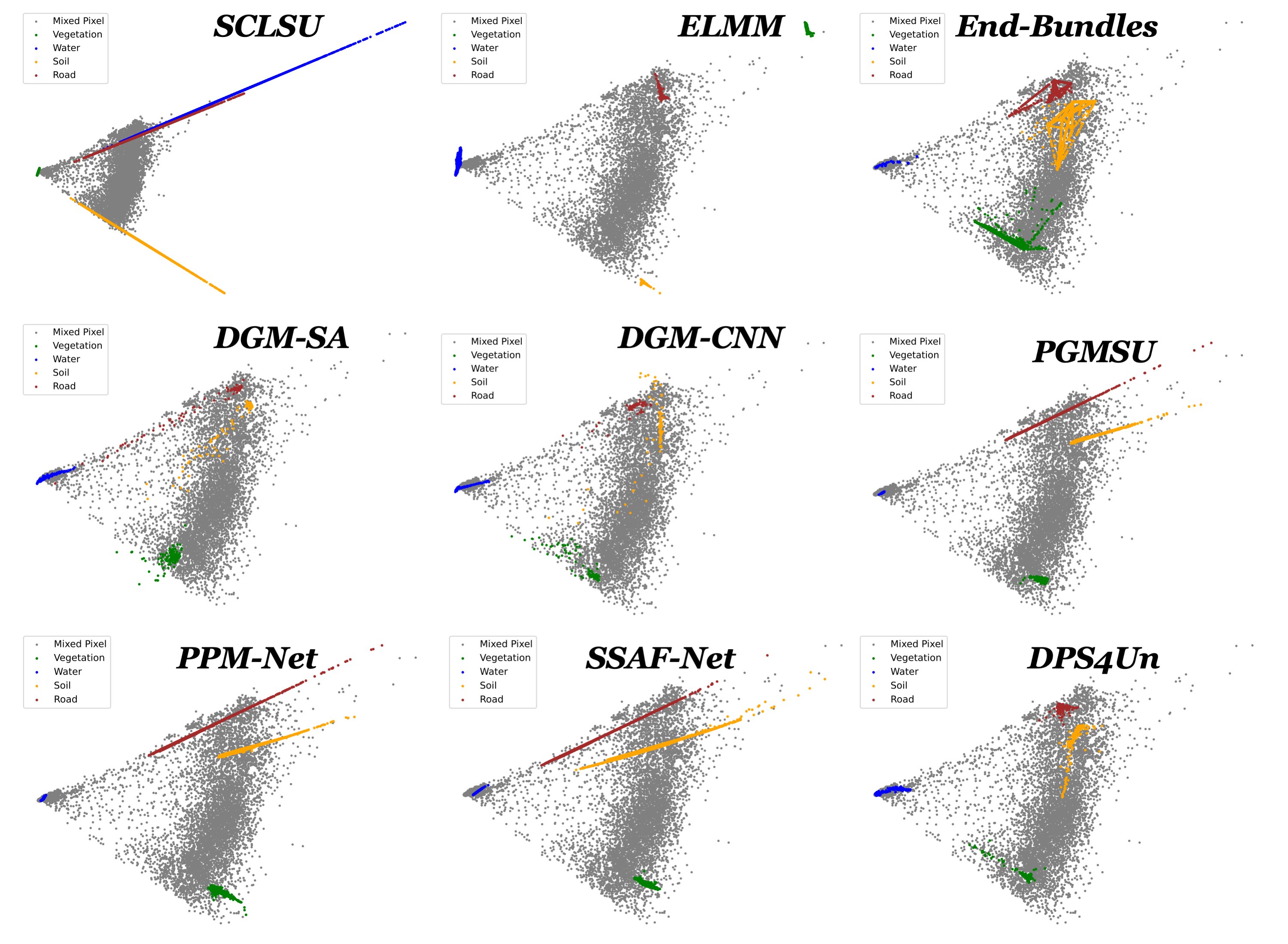}
    \caption{Spectral variability of different methods on Jasper Ridge Dataset.}
    \label{jasper_variability}
\end{figure}
\begin{figure}[!b]
    \centering
    \includegraphics[width=0.49\textwidth]{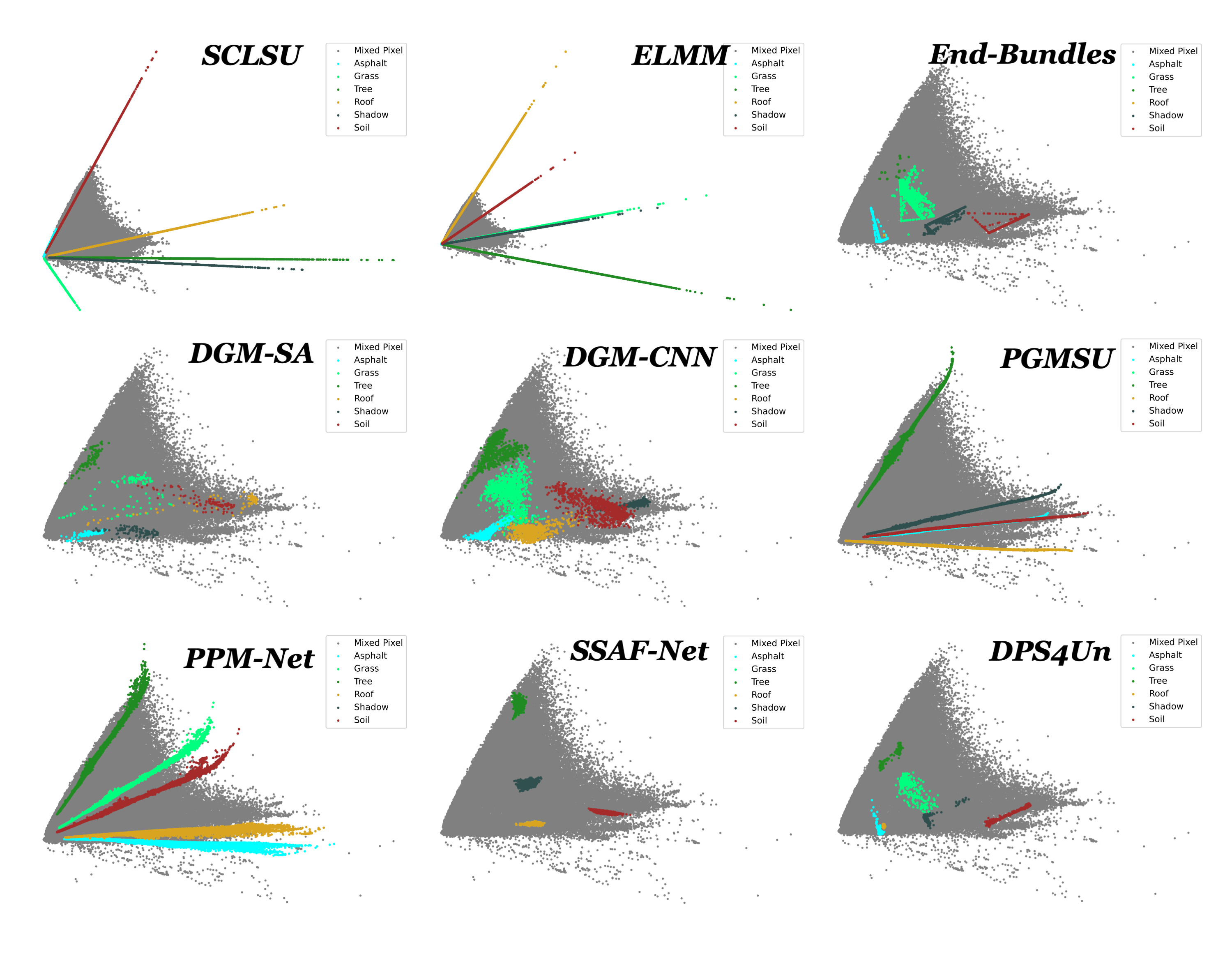}
    \caption{Spectral variability of different methods on Urban Dataset.}
    \label{urban_variability}
\end{figure}

\subsection{Experiments on Real Data}
In this section, we comprehensively evaluate the unmixing
capability of the proposed DPS4Un on both real data and compared with 8 other methods. including SCLSU \cite{drumetz2016blind}, ELMM \cite{8936868}, emdmember bundles \cite{xu2018regional} with FCLSU, DGM-SA \cite{shi2022deep}, DGM-CNN \cite{shi2022deep}, PGMSU \cite{9583297}, PPM-Net \cite{10382701}, SSAF-Net \cite{10896744}, these methods all consider the spectral variability. It should be noticed that endmember boundles is the way we build the image-level spectral library. However, none of these models is a Bayesian inference or spectral prior learning approach. The endmember matching is based on the estimated endmember and the ground truth using the spectral angle distance.
\subsubsection{Results on Jasper Ridge Dataset.}
\autoref{jasper_figures} shows the estimated abundance maps of different methods. As we can see from this figure, SCLSU and ELMM give the wrong estimation. DGM-SA is purely based on self-attention, limiting the capture of the local spatial patterns. Compared with SCLSU, ELMM, DGM-SA and endmember bundles methods, our model gives more detail on soil and road. Similarly, DPS4Un has better abundance estimation performance than endmember bundles with FCLSU, PMM-Net, DGM-Net, and SSAF-Net on vegetation and water. For example, better estimation at the top left part of the vegetation, while the endmember bundles method overestimated the water. For the endmember in \autoref{jasper_curevs}, we do better to compare with the spectral variability modeling approaches. DGM-SA falls short of estimating the endmember, outputting noisy curves, the same with DGM-CNN. SSAF-Net underestimates the roads, and the variability of water is limited. While our model can give more similar results with the ground truth, which has been covered well by the variability. In \autoref{jasper_variability}, we can see that most of the models give linear variability, while our model can better give the non-linear features, which is more reasonable in the real world. \autoref{acc_JR} shows that our model more better at abundance map estimation, achieving a higher aRMSE; all the results are comparable with other state-of-the-art methods.
\begin{figure*}[!t]
    \centering
    \includegraphics[width=\textwidth]{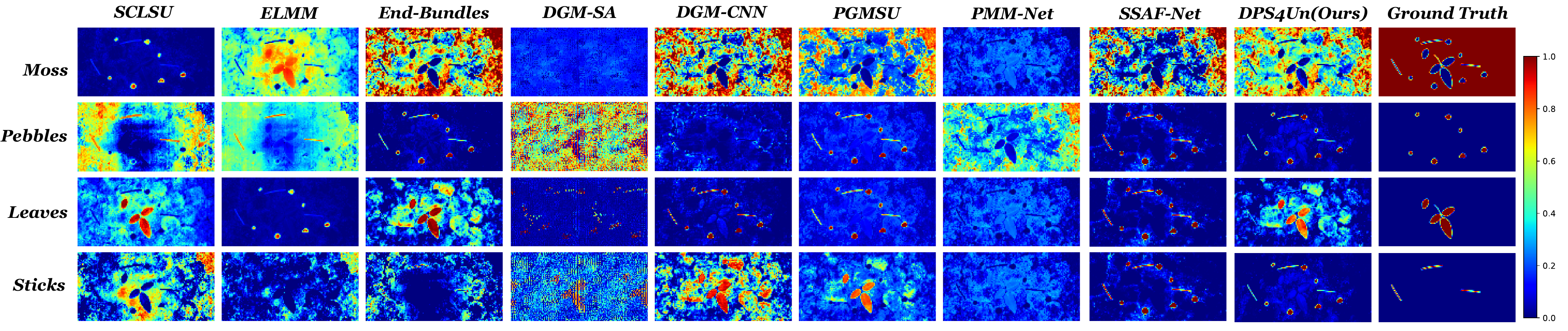}
    \caption{Estimated abundances on the SMScene dataset.}
    \label{RMM_figures}
\end{figure*}

\begin{figure*}[!t]
    \centering
    \includegraphics[width=\textwidth]{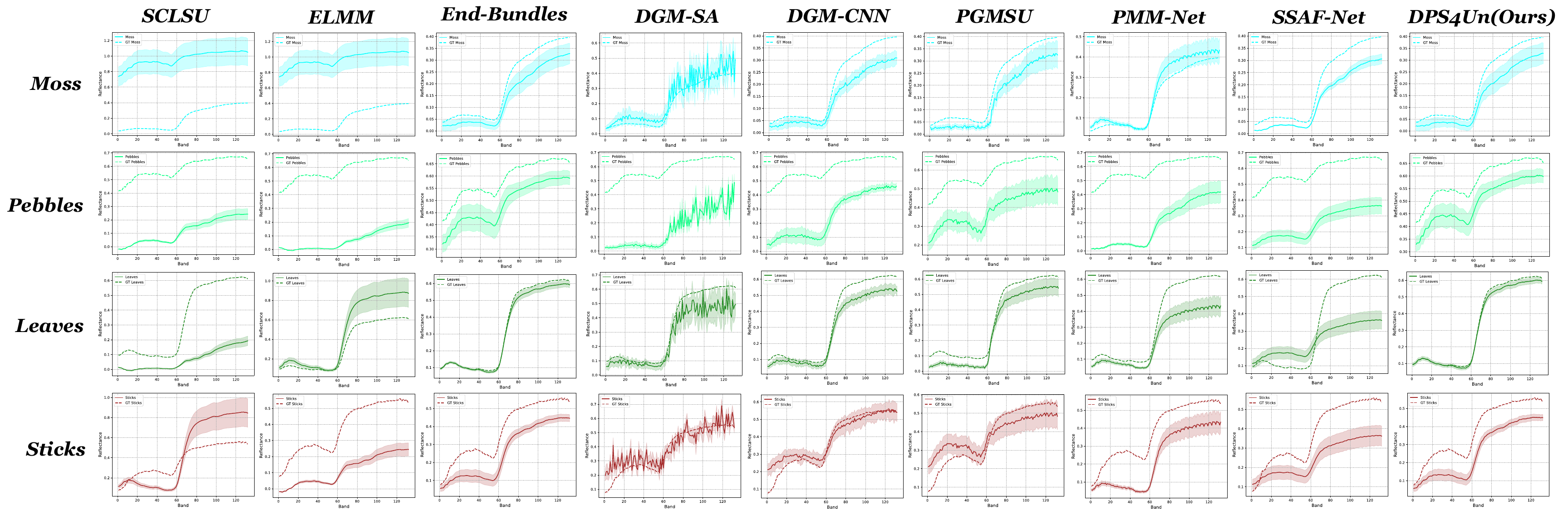}
    \caption{Estimated endmembers on the SMScene dataset. Dash line: ground truth, solid line (or with \(\pm \sigma\) shadow area): estimated endmembers.}
    \label{RMM_curves}
\end{figure*}
\begin{figure}[!h]
    \centering
    \includegraphics[width=0.49\textwidth]{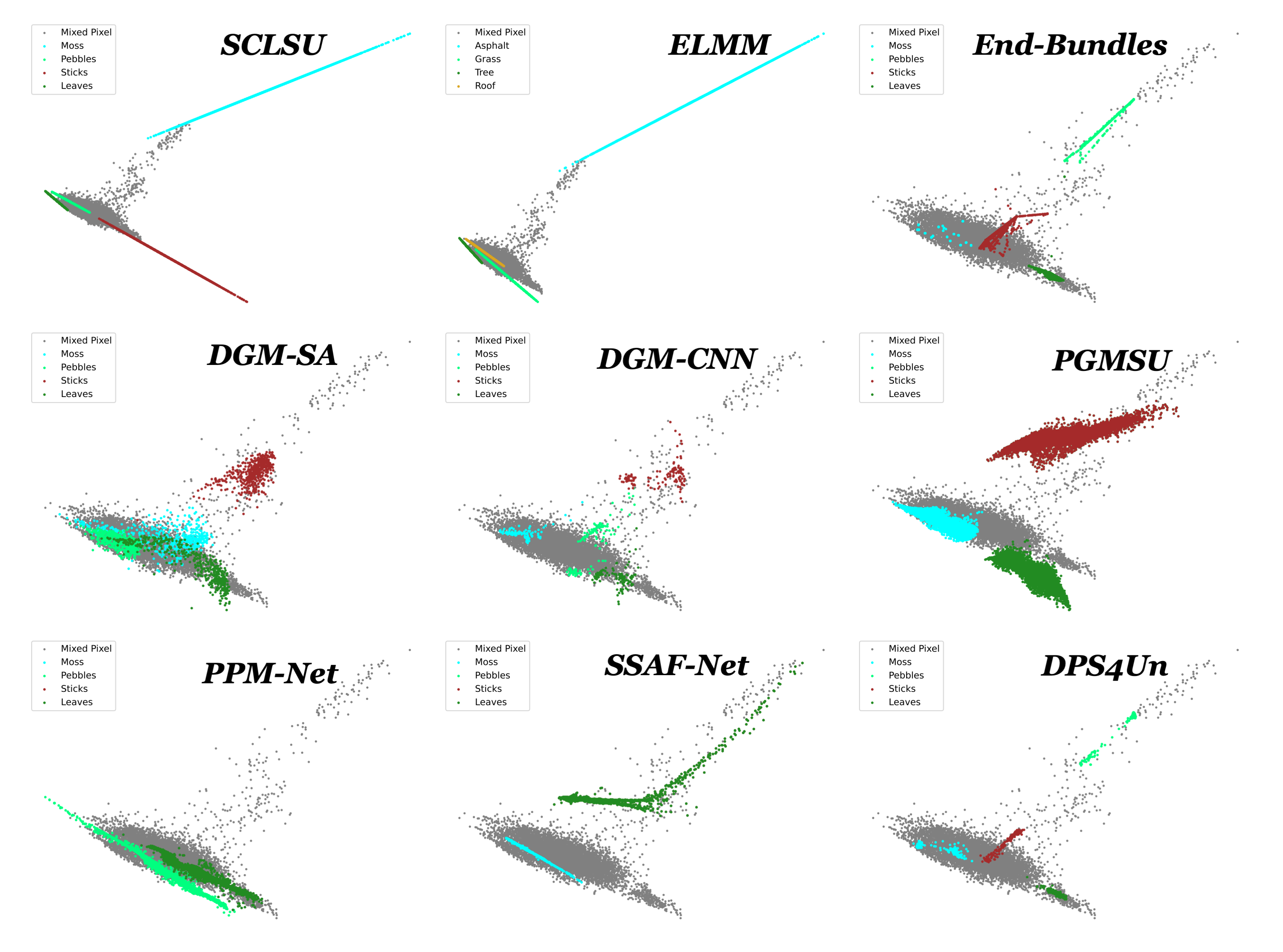}
    \caption{Spectral variability of different methods on SMScene Dataset.}
    \label{RMM_curves}
\end{figure}
\subsubsection{Results on Urban Dataset.}
In \autoref{urban_figures}, our model has better abundance estimation on asphalt, grass, and trees compared with endmember bundles with FCLSU, DGM-SA, DGM-CNN, PMM-Net, and SSAF-Net. Overall, the estimated abundance maps of DPS4Un is more similar to the reference abundance maps. \autoref{acc_urban} shows that DPS4Un is good at endmember estimation, although this dataset scene is complex, and also comparable on abundance RMSE. \autoref{urban_curves} demonstrates that estimated signatures of DPS4Un exhibits high similarity to the reference signatures. DGM-SA, PGMSU, and SSAF-Net have wrong or underestimation on asphalt. The spectral variability distribution of PGMSU, PPM-Net tends to be more linear, and SSAF-Net gives less discrimination on tree and grass, these two are totally overlapping as shown in \autoref{urban_variability}.

\subsubsection{Results on SMScene Dataset.}
\autoref{RMM_figures} shows that our model can better estimate abundance on moss, leaves, and pebbles compared with SCLSU, ELMM, PGMSU, and PMM-Net. SSAF-Net falls short in leaf finding. For the spectral variability modeling, SCLSU and ELMM give extreme values of endmembers, showing outlier features. Our model demonstrates more discriminative capability on spectral variability modeling.

\subsubsection{Visualization of Sampling Trajectory}
\begin{figure}[!h]
    \centering
    \includegraphics[width=0.49\textwidth]{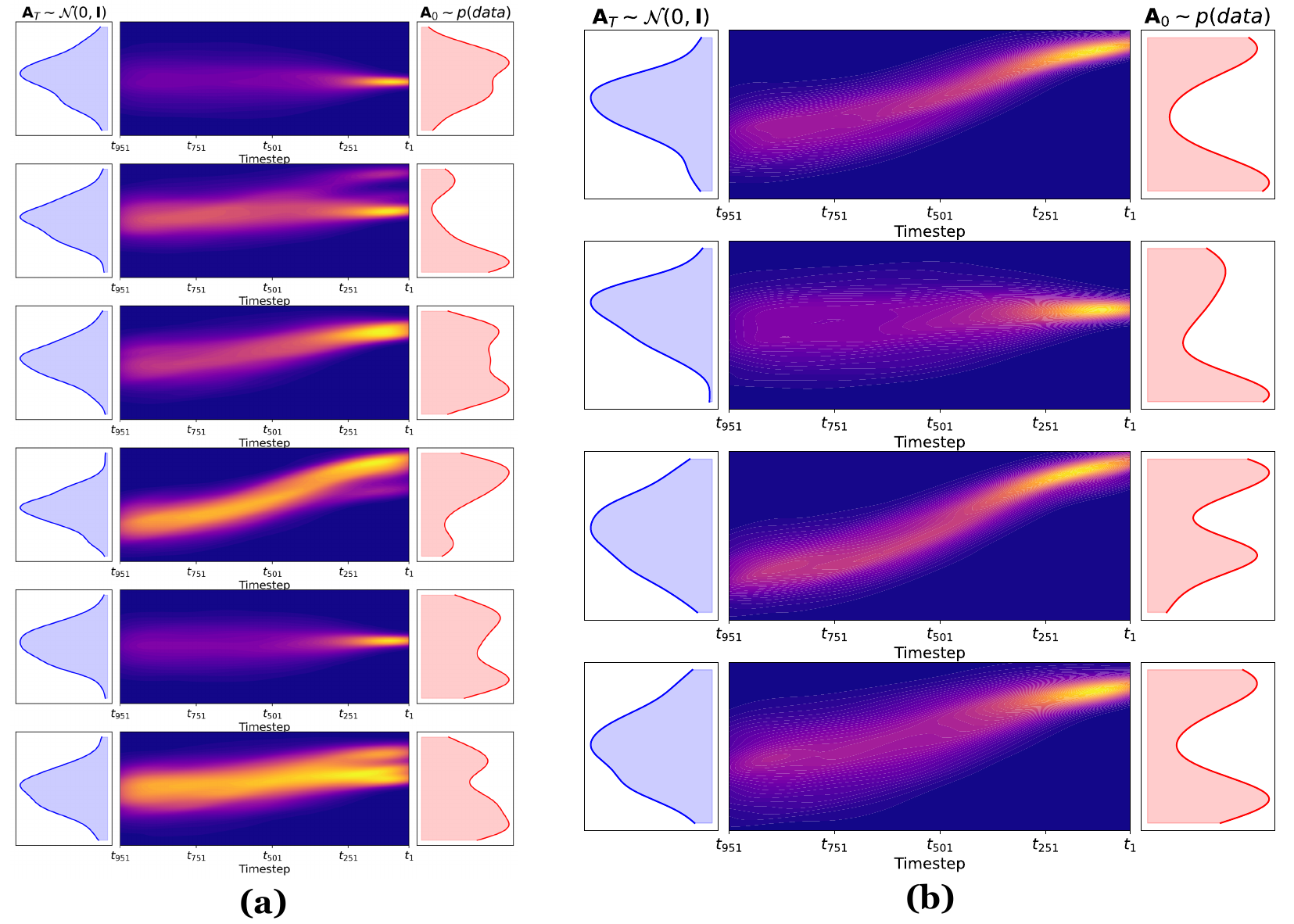}
    \caption{Sampling trajectory of two datasets. The heatmap shows the density of the trajectory, where the number of trajectories is defined by the number of superpixels. (a) Urban, (d) SMScene.}
    \label{trajectory}
\end{figure}

Interestingly, we visualize the sampling path of our DPS4Un model. In our model, we assign pure endmember sets within each superpixel, meaning that each superpixel (or homogenous area) should have its own endmembers, contributing to spectral variability modeling. By performing Bayesian inference, our model can generate data from the Gaussian noise distribution to the desired data/endmember distribution, as we can see from \autoref{figure1} and \autoref{trajectory}. Overall, the reverse process flow toward high probability regions of the data distribution, showing that our proposed DPS4Un can path directly to the high-mass target region, which is good for the spectral unmixing problem.

\begin{table}[!h]
\caption{Ablation study on the gradient updating step of abundance estimation.}
\resizebox{0.49\textwidth}{!}{
\begin{tabular}{c|ccccc}
\hline
                         & \(\lambda\) & 1 & 0.1 & 0.01 & 0.001 \\ \hline
\multirow{2}{*}{Jasper Ridge}      & aRMSE  & 0.2478 & \colorbox{mycolor}{\textbf{0.0517}} & 0.0668 & 0.0624 \\
                         & aSAD   & 0.0888 & \colorbox{mycolor}{\textbf{0.0785}} & 0.1054 & 0.0897 \\ \hline
\multirow{2}{*}{Urban}   & aRMSE  & 0.3073 & \colorbox{mycolor}{\textbf{0.1751}} & 0.1769 & 0.2399 \\
                         & aSAD   & 0.0984 & 0.0997 & \colorbox{mycolor}{\textbf{0.0975}} & 0.0987 \\ \hline
\multirow{2}{*}{SMScene} & aRMSE  & 0.4026 & \colorbox{mycolor}{\textbf{0.2098}} & 0.2463 & 0.3115 \\
                         & aSAD   & 0.0717 & 0.0704 & \colorbox{mycolor}{\textbf{0.0685}} & 0.0695 \\ \hline
\end{tabular}
}
\label{ablation study}
\end{table}
\subsubsection{Parameter sensitivity on \(\lambda\)}
In Algorithm \autoref{algo}, we have parameter \(\lambda\) that control the intensity of gradient of \(\nabla_{\boldsymbol{A}_t} \log p(\hat{\boldsymbol{S}_0} | \hat{\boldsymbol{A}_0})\). From \autoref{ablation study}, we can see that when \(\lambda\) equals 0.1, the DPS4Un provides a more precise estimation of abundance maps and endmembers.


%% file: 5_Conclusion.tex
\section{Conclusion} \label{sec:conclusion}

Standing at the forefront of inverse problems in hyperspectral unmixing, we extend the diffusion model and propose a novel Bayesian inference paradigm for solving the spectral unmixing problem in HSI. Traditionally, deep learning based methods are performed on the high-dimensional HSI cube space using AE-like architecture for semi-blind unmixing. While, our proposed method DPS4Un shifts to a more manageable, lower-dimensional space. Additionally, we rethink the spectral prior learning from the perspective of the cutting-edge generative diffusion model, instead of using VAE, which may lead to posterior collapse. This ability allows to represent more diverse characteristics of the spectrum and captures the spectral variability, and generalizes to real-world data. The experimental results show that DPS4Un has comparable and better performance on abundance estimation, and the sampling trajectories demonstrate that the flow of gradient is more reliable towards the ideal endmember distribution.